%% file: acl_latex.tex
\pdfoutput=1

\documentclass[11pt]{article}

\usepackage[preprint]{acl}

\usepackage{times}
\usepackage{multirow}
\usepackage{latexsym}
\usepackage{enumitem}
\usepackage{amsmath}
\usepackage{tikz}
\usepackage{graphicx}
\usepackage{colortbl}
\usepackage{array}
\usepackage{booktabs}
\usetikzlibrary{positioning}
\usetikzlibrary{shapes.callouts, shapes.geometric, arrows}
\usepackage{todonotes}
\usepackage{hyperref}
\usepackage{subcaption}
\usepackage[T1]{fontenc}

\usepackage[utf8]{inputenc}

\usepackage{microtype}

\usepackage{inconsolata}

\usepackage{graphicx}

\newcommand{\thickcline}[1]{%
  \noalign{\global\arrayrulewidth=0.7pt}%
  \cline{#1}%
  \noalign{\global\arrayrulewidth=0.4pt}%
}

\title{Liar, Liar, Logical Mire: A Benchmark for Suppositional Reasoning\\ in Large Language Models}

\author{Philipp Mondorf\textsuperscript{\normalfont 1, 2} \and
         Barbara Plank\textsuperscript{\normalfont 1, 2}\\
  \textrm{\textsuperscript{1}}MaiNLP, Center for Information and Language Processing, LMU Munich, Germany \\
  \textrm{\textsuperscript{2}}Munich Center for Machine Learning (MCML), Munich, Germany \\
{\tt p.mondorf@lmu.de} \hspace{2em} {\tt b.plank@lmu.de}}

\begin{document}
\maketitle
\begin{abstract}
Knights and knaves problems represent a classic genre of logical puzzles where characters either tell the truth or lie. The objective is to logically deduce each character's identity based on their statements. The challenge arises from the truth-telling or lying behavior, which influences the logical implications of each statement. Solving these puzzles requires not only direct deductions from individual statements, but the ability to assess the truthfulness of statements by reasoning through various hypothetical scenarios. As such, knights and knaves puzzles serve as compelling examples of suppositional reasoning. In this paper, we introduce \emph{TruthQuest}, a benchmark for suppositional reasoning based on the principles of knights and knaves puzzles. Our benchmark presents problems of varying complexity, considering both the number of characters and the types of logical statements involved. Evaluations on \emph{TruthQuest} show that large language models like Llama 3 and Mixtral-8x7B exhibit significant difficulties solving these tasks. A detailed error analysis of the models' output reveals that lower-performing models exhibit a diverse range of reasoning errors, frequently failing to grasp the concept of truth and lies. In comparison, more proficient models primarily struggle with accurately inferring the logical implications of potentially false statements.
\end{abstract}

\section{Introduction}\label{sec:introduction}
\begin{figure}[t!]
  \centering 
  \input{tikz/introduction/intro_problem}
  \caption{
  An instance of the \emph{knights \& knaves} puzzle. By reasoning about the characters' statements and their truthfulness, it is possible to deduce that Greeny and Bluey must be knights, while Pinky is a knave.
  }
  \label{fig:knights_and_knaves_example_1}
\end{figure}

Well-designed logic puzzles can serve as a valuable tool for gaining deeper insights into the capabilities of large language models (LLMs)~\cite{giadikiaroglou2024puzzle, li2024assessing, del-fishel-2023-true}. By challenging models to navigate sophisticated logic problems, these puzzles can reveal how LLMs identify patterns, recognize relationships, and employ logical principles \cite{tong2023eliminating, ding2024thoughts}. In his book \emph{"What is the Name of This Book?"}, \citet{smullyan1978whatis} introduced a series of \emph{knights and knaves} puzzles, where characters are either knights who always tell the truth or knaves who always lie.\footnote{Note:\ puzzles of this kind have existed before under different variations and names \cite{maurice1953mathematical, goodman1972problems}.} The goal is to deduce the identity of each character based on their statements (see Figure \ref{fig:knights_and_knaves_example_1}). Unlike other deductive reasoning tasks, where premises are typically assumed to be true \cite{han2024folio, dalvi-etal-2021-explaining, 10.5555/3491440.3491977}, these puzzles require the reasoner to assess the truthfulness of statements by exploring different hypothetical scenarios. For instance, if the statement of \emph{Pinky} in Figure \ref{fig:knights_and_knaves_example_1} were true, \emph{Greeny} must be a knight, thus telling the truth. However, \emph{Greeny} states that \emph{Pinky} is lying, which contradicts the initial truth assumption of \emph{Pinky}'s statement. Hence, \emph{Pinky} must be a knave. If \emph{Pinky} is a knave, then \emph{Greeny}'s statement is true, and thus \emph{Greeny} will be a knight. Based on \emph{Pinky}'s false statement, it then follows that \emph{Bluey} must also be a knight. This form of suppositional reasoning, i.e.\ the ability to reason conditionally, is essential in scenarios where the logical ramifications of different possibilities need to be considered, such as in planning or everyday reasoning \cite{BYRNE19971}.

In this paper, we introduce \emph{TruthQuest},\footnote{\href{https://huggingface.co/datasets/mainlp/TruthQuest}{https://huggingface.co/datasets/mainlp/TruthQuest}} a benchmark designed to evaluate the suppositional reasoning capabilities of large language models through knights and knaves puzzles. We present 2,400 problems of varying complexity, depending on the number of characters and types of logical statements involved (see Section \ref{sec:dataset}). We assess the reasoning behavior of three model families: Llama 2~\cite{touvron2023llama}, Llama 3~\cite{meta2024llama3}, and Mixtral-8x7B~\cite{mistral2023mixtral}. In addition to evaluating the models' task performance, we conduct an in-depth analysis of their outputs to gain insights into the types of errors encountered during reasoning. This is done through both comprehensive human inspection and AI-assisted evaluation. Our findings reveal that:

\begin{itemize}[itemsep=0pt, topsep=2pt]
    \item All models exhibit significant difficulties in solving knights and knaves problems.
    \item Although more advanced prompting techniques, such as chain-of-thought prompting \cite{NEURIPS2022_9d560961}, enhance performance on simpler problems, accuracy declines markedly as puzzle complexity increases.
    \item The types of reasoning errors are closely linked to the models' performance. Lower-performing models exhibit a diverse range of reasoning errors, while more proficient models primarily struggle with deducing the correct logical implications of false statements.
\end{itemize}

\section{Related Work}\label{sec:related_work}

\paragraph{Deductive Reasoning with LLMs.} Several studies evaluate LLMs on deductive reasoning tasks~\cite{saparov2023language, NEURIPS2023_deb3c281, wan2024}. In line with our research, some of these works employ logical puzzles to analyze the reasoning behaviors of LLMs~\cite{7aceacfcd0f44c7cb19a9e3d7181e038, NEURIPS2023_271db992, Jiang2023BRAINTEASERLT}. However, to the best of our knowledge, \emph{TruthQuest} is the first deductive reasoning benchmark that evaluates the ability of LLMs to infer both the truthfulness of statements and their logical implications.

\begin{table*}[thb]
\centering
{\fontsize{8}{10}\selectfont
\begin{tabular}{>{\centering\arraybackslash}p{0.125\textwidth} p{0.42\textwidth} p{0.25\textwidth} p{0.0\textwidth} p{0.01\textwidth} p{0.01\textwidth} p{0.01\textwidth}}
\toprule
\textbf{Statement Types} & \textbf{Natural Language Example} & \textbf{Logic Expression} & & \multicolumn{3}{c}{\textbf{Set}} \\
\specialrule{1pt}{0pt}{0pt}
Self-Referential & $P_i$: \emph{I am a knight} & $P_i$ & & \cellcolor{cyan!25} & & \rule{0pt}{3ex}\\
\thickcline{1-4}
Accusation & $P_i$: \emph{$P_j$ is a knight/knave} & $P_i \Leftrightarrow \psi_j \; (i \neq j)$ & & \cellcolor{cyan!25} & \cellcolor{yellow!25} & \cellcolor{green!25} \rule{0pt}{3ex} \\
\thickcline{1-4}
Conjunction & $P_i$: \emph{$P_j$ is a knight/knave and $P_k$ is a knight/knave} & $P_i \Leftrightarrow \left(\psi_j \land \psi_k\right) \; \, (i \neq j \neq k)$ & & \multirow{-4}{*}{S \cellcolor{cyan!25}} & \cellcolor{yellow!25} & \cellcolor{green!25} \rule{0pt}{3ex} \\
\thickcline{1-5}
Implication & $P_i$: \emph{If $P_j$ is a knight/knave, then $P_k$ is a knight/knave} & $P_i \Leftrightarrow \left(\psi_j \rightarrow \psi_k\right) \, (i \neq j \neq k)$ & & &  \multirow{-4}{*}{I \cellcolor{yellow!25}} & \cellcolor{green!25} \rule{0pt}{3ex} \\
\thickcline{1-6}
Equivalence & $P_i$: \emph{$P_j$ is a knight/knave if and only if $P_k$ is a knight/knave} & $P_i \Leftrightarrow \left(\psi_j \Leftrightarrow \psi_k\right) \, (i \neq j \neq k)$ & & & \rule{0pt}{3ex} & \multirow{-5}{*}{E \cellcolor{green!25}}\\
\specialrule{1pt}{0pt}{0pt}
\end{tabular}
}
\caption{Character statements in \emph{TruthQuest}. Each type is represented by an example expressed both in natural language and boolean logic. The final column indicates the types of statements included in each statement set. For instance, $S$ is the only set that includes self-referential statements alongside accusations and conjunctions.}
\label{tab:statement_types}
\end{table*}

\section{Dataset}\label{sec:dataset}
Various versions of knights and knaves puzzles exist \cite{smullyan1978whatis, JOHNSONLAIRD199069}. In this work, we focus on the most popular variant, which features only two types of characters: knights, who always tell the truth, and knaves, who always lie, as illustrated in Figure \ref{fig:knights_and_knaves_example_1}. To construct valid instances of knights and knaves problems, we formalize the puzzle using a two-valued logic. Specifically, knights are assigned the truth value \emph{true}, while knaves are mapped to \emph{false}. For a given puzzle with $n$ characters, where $P$ denotes the truth value of a character and $Q$ is the character's logical claim, the puzzle can be expressed as a single conjunction using the bi-conditional operator:

\begin{equation}\label{eq:puzzle_expression}
    \begin{aligned}
    \Phi  = & \left(P_1 \Leftrightarrow Q_1\right) \land \left(P_2 \Leftrightarrow Q_2\right) \\
    & \land \ldots \land \left(P_n \Leftrightarrow Q_n\right) 
\end{aligned}
\end{equation}

For instance, the example depicted in Figure \ref{fig:knights_and_knaves_example_1} can be expressed as:
\begin{align*}
    \Phi & = \left(P_1 \Leftrightarrow \neg P_3\right) \land P_2 \land \left(P_3 \Leftrightarrow \left(P_1 \, \land \, \neg P_2\right)\right)
\end{align*}
where $P_1$, $P_2$, and $P_3$ correspond to the truth values of Greeny, Bluey, and Pinky, respectively. In this example, the characters' statements are given as follows: $Q_1 = \neg P_3$ (``Pinky is a knave!''), $Q_2 = P_2$ (``I am a knight!''), and $Q_3 = \left(P_1 \, \land \, \neg P_2\right)$ (``Greeny is a knight and Bluey is a knave!''). The bi-conditional operator ($P_i \Leftrightarrow Q_i$) reflects that a character is either lying or telling the truth. Specifically, it indicates that the statement $Q_i$ holds true if and only if the character is a knight (denoted by $P_i$). Conversely, if the character is a knave, $\neg P_i$, then the corresponding statement is false, $\neg Q_i$. To derive all $m$ possible solutions for such a puzzle, the expression $\Phi$ can be transformed into disjunctive normal form using Boolean algebra:

\begin{equation}\label{eq:puzzle_solution}
\begin{aligned}
    \Phi = &\left(\psi_1^1 \land \ldots \land \psi_n^1\right) \, \lor \, \left(\psi_1^2 \land \ldots \land \psi_n^2\right) \\ 
    &\lor \, \ldots \, \lor \, \left(\psi_1^m  \land \ldots \land \psi_n^m\right)
\end{aligned}
\end{equation}

where $\psi_i^j \in \{P_i, \neg P_i\}$ denotes the character's identity as either knight ($P_i$) or knave ($\neg P_i$).

\paragraph{Dataset Creation.} For \emph{TruthQuest}, we limit character statements to the types outlined in Table \ref{tab:statement_types}. To examine the impact of statement types on model behavior, we classify them into three distinct sets: $S$, $I$, and $E$, as specified in the table. For each set, separate datasets of knights and knaves puzzles are generated. Specifically, instances are created by randomly sampling the statement of each character from the respective set, $Q_i \sim C \in \{S, I, E\}$. Each puzzle is solved by converting the problem (Equation \ref{eq:puzzle_expression}) into disjunctive normal form, as shown in Equation \ref{eq:puzzle_solution}.  While Boolean algebra is used to construct the dataset, all statements and solutions are ultimately presented in natural language. Each dataset sample comprises the characters' statements and the corresponding solution, representing the logically valid identity of each character (as illustrated in Figure \ref{fig:knights_and_knaves_example_1}). For our benchmark, we include only instances that have a single, unique solution, i.e., $m = 1$. Furthermore, we consider varying numbers of characters for each statement set, specifically: $n = {3, 4, 5, 6}$. This yields $3 \times 4 = 12$ data subsets. For each subset, 200 problems are generated, resulting in a total of 2,400 unique instances.

\section{Experimental Setup}\label{sec:exp_setup}

\paragraph{Language Models.} We assess a total of six LLMs from three prominent open-access model families: Llama 2 (7B, 13B and 70B), Llama 3 (8B and 70B), and Mixtral-8x7B. The publicly accessible weights are obtained from the Hugging Face platform, specifically \texttt{Llama-2-chat-hf},\footnotemark[3] \texttt{Meta-Llama-3-Instruct},\footnote{\href{https://huggingface.co/meta-llama}{huggingface.co/meta-llama}} and \texttt{Mixtral-8x7B-Instruct-v0.1}.\footnote{\href{https://huggingface.co/mistralai/Mixtral-8x7B-Instruct-v0.1}{huggingface.co/mistralai/Mixtral-8x77B-Instruct-v0.1}} For further details about the models and prompts we employ, please refer to Appendix \ref{appendix:language_models}.

\paragraph{Evaluation Framework.} To assess the models' task performance, we follow a two-step approach. First, we use regular expressions to parse the models' final conclusions according to the format specified in the input prompt. If responses cannot be parsed in this way, they are subsequently passed to an additional language model, specifically LLaMA-3-8B, which extracts the conclusion in the desired format (for the full evaluator prompt, see Figure \ref{fig:conclusion_evaluator_prompt} in the appendix). A schematic overview of this approach is presented in Figure \ref{fig:conclusion_evaluator} in the appendix. Once a conclusion is successfully parsed, the models' accuracy is determined by an exact match between their prediction and the correct puzzle solution. For a prediction to be considered accurate, the models must correctly deduce the identity of all characters.

Beyond assessing task performance, we analyze the models' reasoning errors. We manually inspect a subset of the responses from LLaMA-3-8B (zero-shot) and LLaMA-3-70B (zero-shot and four-shot chain-of-thought prompting). Specifically, we evaluate 10 responses from each of the 12 data subsets for each model and setup, totaling 360 responses. This involves parsing the model's conclusion and assessing its reasoning against six common error categories previously devised, as outlined in Table \ref{tab:error_descriptions} in the appendix. This comprehensive manual evaluation is conducted independently by two hired students with expertise in data annotation. To assess the quality of the annotations, we report an overall Cohen’s Kappa value of $\kappa = 0.70$. For a detailed description of the manual evaluation procedure and an overview of the inter-annotator agreement for each error type, please refer to Section \ref{appendix:human_evaluation} in the appendix. To complement our manual evaluation and assess all model responses with respect to the error categories devised, we leverage GPT-4 \cite{openai2024gpt4} using few-shot prompting. For the complete prompt with all few-shot examples, see Figures \ref{fig:error_evaluator_prompt} to \ref{fig:example_6_error_evaluator_prompt} in the appendix.

\begin{table*}[thp]
{\fontsize{8}{10}\selectfont
\centering
\begin{tabular}{c|c|cccc|cccc|cccc}
\toprule
\multirow{2}{*}{\textbf{Model}} & \multirow{2}{*}{\textbf{Mode}} & \multicolumn{4}{c|}{\textbf{Set S}} & \multicolumn{4}{c|}{\textbf{Set I}} & \multicolumn{4}{c}{\textbf{Set E}} \\
\cmidrule(lr){3-6} \cmidrule(lr){7-10} \cmidrule(lr){11-14}
& & \textbf{3} & \textbf{4} & \textbf{5} & \textbf{6} & \textbf{3} & \textbf{4} & \textbf{5} & \textbf{6} & \textbf{3} & \textbf{4} & \textbf{5} & \textbf{6} \\
\midrule
\textbf{Random Baseline} & - & 0.13 & 0.06 & 0.03 & 0.02 & 0.13 & 0.06 & 0.03 & 0.02 & 0.13 & 0.06 & 0.03 & 0.02 \\
\midrule
\textbf{LLaMA-2-7b} & \multirow{6}{*}{\textbf{zero shot}} & 0.08 & 0.06 & 0.02 & 0.00 & 0.20 & 0.10 & 0.07 & 0.04 & 0.21 & 0.11 & 0.03 & 0.03 \\
\textbf{LLaMA-2-13b} & & 0.10 & 0.06 & 0.03 & 0.03 & 0.13 & 0.11 & 0.04 & 0.02 & 0.15 & 0.08 & 0.05 & 0.01 \\
\textbf{LLaMA-2-70b} & & 0.13 & 0.13 & 0.08 & 0.03 & 0.13 & 0.11 & 0.06 & 0.03 & 0.17 & 0.09 & 0.07 & 0.02 \\
\textbf{LLaMA-3-8B} & & 0.07 & 0.13 & 0.04 & 0.04 & 0.19 & \textbf{0.18} & 0.07 & 0.04 & 0.13 & 0.08 & 0.06 & 0.03 \\
\textbf{LLaMA-3-70B} & & \textbf{0.29} & \textbf{0.22} & \textbf{0.13} & \textbf{0.10} & \textbf{0.32} & 0.14 & \textbf{0.14} & \textbf{0.11} & \textbf{0.29} & \textbf{0.18} & \textbf{0.11} & \textbf{0.06} \\
\textbf{Mixtral-8x7B} & & 0.16 & 0.08 & 0.04 & 0.03 & 0.21 & 0.14 & 0.06 & 0.05 & 0.17 & 0.08 & 0.04 & 0.01 \\
\midrule
\multirow{5}{*}{\textbf{LLaMA-3-70B}} & \textbf{four shot} & 0.22 & 0.25 & 0.19 & 0.13 & 0.24 & 0.21 & 0.13 & 0.10 & 0.32 & 0.22 & 0.11 & 0.05 \\
& \textbf{eight shot} & 0.22 & 0.21 & 0.16 & 0.09 & 0.32 & 0.25 & 0.07 & 0.09 & 0.27 & 0.20 & 0.10 & 0.02 \\
& \textbf{zero CoT} & 0.23 & 0.17 & 0.14 & 0.12 & 0.28 & 0.17 & 0.15 & 0.09 & 0.29 & 0.17 & \textbf{0.12} & 0.08 \\
& \textbf{four CoT} & 0.46 & \textbf{0.31} & \textbf{0.21} & 0.16 & \textbf{0.33} & \textbf{0.27} & 0.11 & \textbf{0.15} & \textbf{0.40} & \textbf{0.25} & \textbf{0.12} & \textbf{0.10} \\
& \textbf{eight CoT} & \textbf{0.60} & 0.26 & \textbf{0.21} & \textbf{0.20} & \textbf{0.33} & 0.20 & \textbf{0.15} & 0.12 & 0.37 & 0.20 & \textbf{0.12} & \textbf{0.10} \\
\bottomrule
\end{tabular}
}
\caption{Accuracy values for different models and prompting techniques across each subset of \emph{TruthQuest}. Results are grouped first by prompting technique and then by model. Bold values represent highest performance among a group. The random baseline indicates the accuracy achieved by guessing the identity of each character.}
\label{tab:accuracies}
\end{table*}

\paragraph{Meta-Evaluation.} We assess the quality of our evaluation procedures by comparing the results obtained via automatic evaluation with our manual assessment. Respective results are reported in Section \ref{sec:results} and Appendix \ref{appendix:ai_assisted_evaluation}.

\section{Results}\label{sec:results}

\paragraph{Task Performance}\label{subsec:task_performance}
Table \ref{tab:accuracies} provides an overview of the models' task performance on \emph{TruthQuest}. The table includes results for all models when prompted in a zero-shot setting, with additional results for LLaMA-3-70B using various prompting techniques (detailed results for other models can be found in Table \ref{tab:full_accuracy_values} in the appendix). We observe that under zero-shot prompting, all models exhibit relatively poor performance across the different data subsets, often performing at or below chance level. 

Although LLaMA-3-70B generally outperforms other models, its accuracy significantly declines as the number of characters—and consequently, the number of inference steps—increases. When guided via chain-of-thought prompting, LLaMA-3-70B shows performance improvements for problems involving fewer characters, particularly with statements sampled from set $S$. Other prompting techniques, such as few-shot prompting or zero-shot CoT \cite{NEURIPS2022_8bb0d291}, do not substantially enhance LLaMA-3-70B's task performance.

\paragraph{Content Effects.} For our analysis, we replace the terms \emph{knights} and \emph{knaves} with the pseudo-words \emph{jabbas} and \emph{tettes} to reduce the likelihood that models have been exposed to similar problems during training. Interestingly, we find that the choice of terms for \emph{knights} and \emph{knaves} seems to have no substantial impact on the models' performance, as shown in Figure \ref{fig:content_effects} in the appendix.

\begin{figure}[b!]
    \centering
    \begin{subfigure}[b]{0.39\textwidth}
        \centering
        \includegraphics[width=\linewidth]{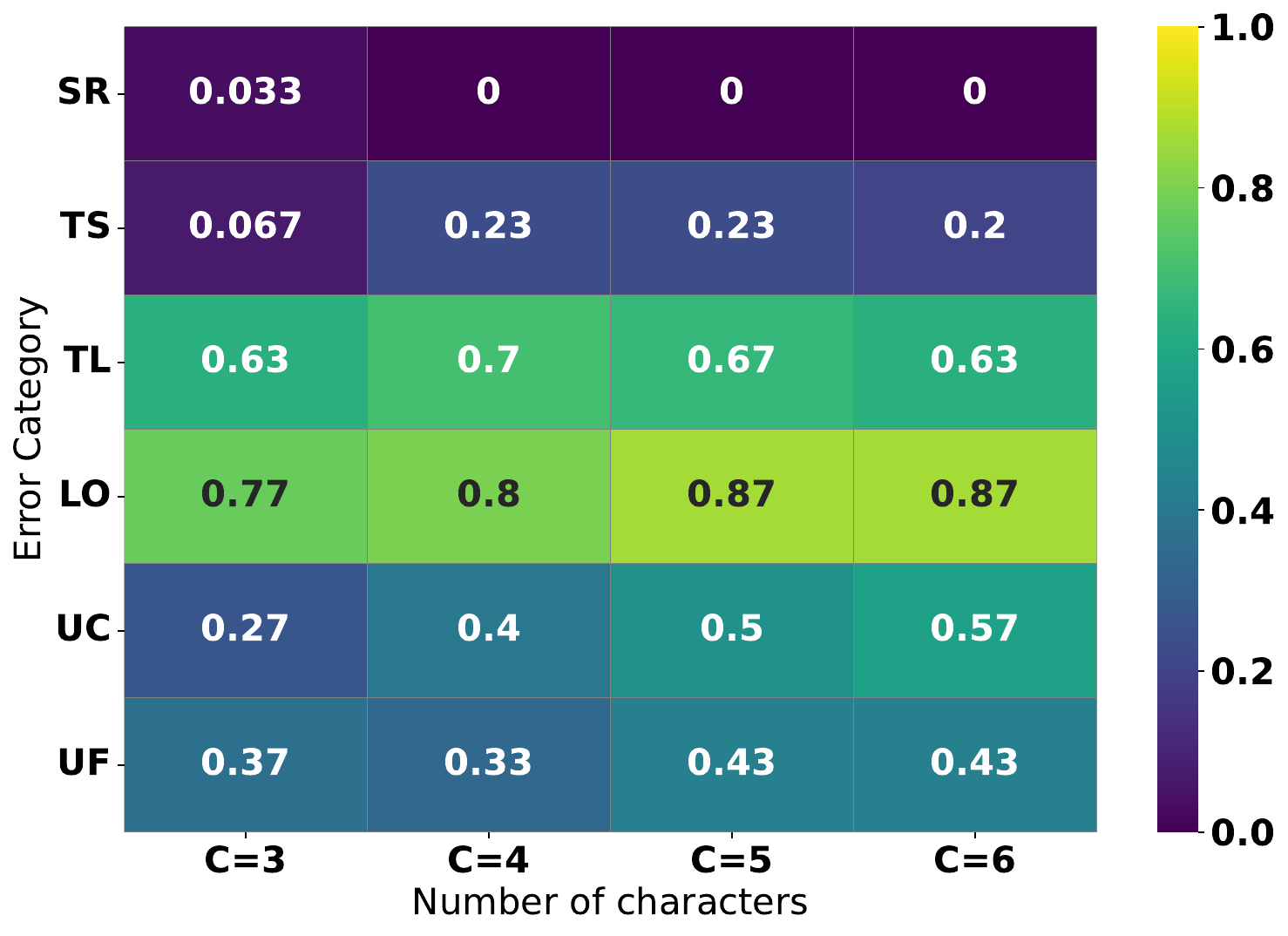}
        \caption{LLaMA-3-8B (zero-shot)}
        \label{fig:llama3_8b_zero_shot_human_reasoning_eval}
    \end{subfigure}\\[0.5em]
    \begin{subfigure}[b]{0.39\textwidth}
        \centering
        \includegraphics[width=\linewidth]{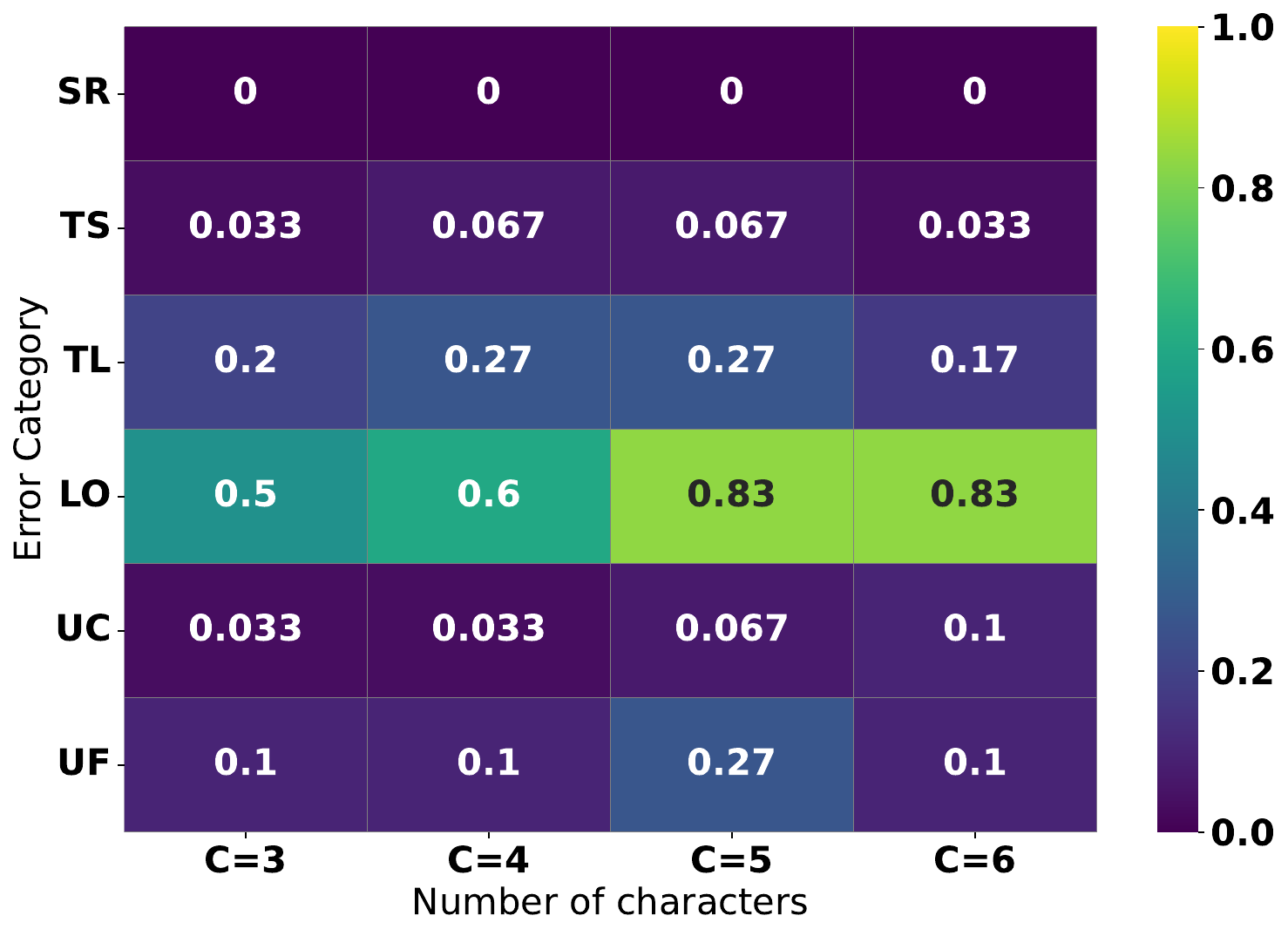}
        \caption{LLaMA-3-70B (four-shot CoT)}
        \label{fig:llama3_70b_four_cot_human_reasoning_eval}
    \end{subfigure}
    \caption{Relative occurrence of reasoning errors. Values are obtained from human-based evaluations. Error categories are abbreviated: \emph{(SR) False statement reproduction}, \emph{(TS) Assuming statements to be true}, \emph{(TL) Misunderstanding the concept of truth and lies}, \emph{(LO) Misunderstanding logical operators}, \emph{(UC) Unjustified conclusion}, and \emph{(UF) Unfaithfulness} (see Table \ref{tab:error_descriptions}).}
    \label{fig:human_error_analysis}
\end{figure}

\begin{figure}[b!]
    \centering
    \begin{subfigure}[b]{0.39\textwidth}
        \centering
        \includegraphics[width=\linewidth]{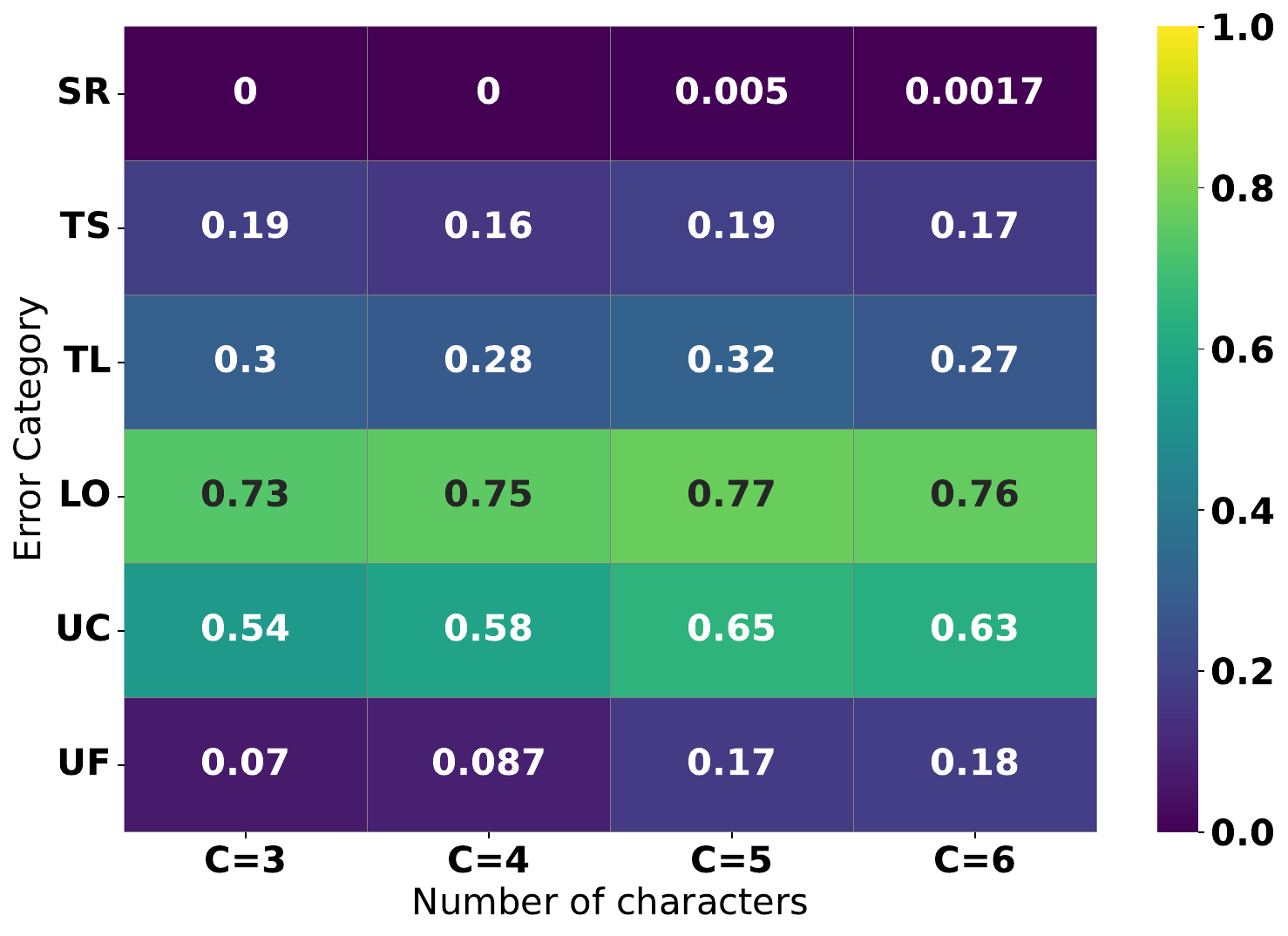}
        \caption{LLaMA-3-8B (zero-shot)}
        \label{fig:llama3_8b_zero_shot_gpt4_reasoning_eval}
    \end{subfigure}\\[0.5em]
    \begin{subfigure}[b]{0.39\textwidth}
        \centering
        \includegraphics[width=\linewidth]{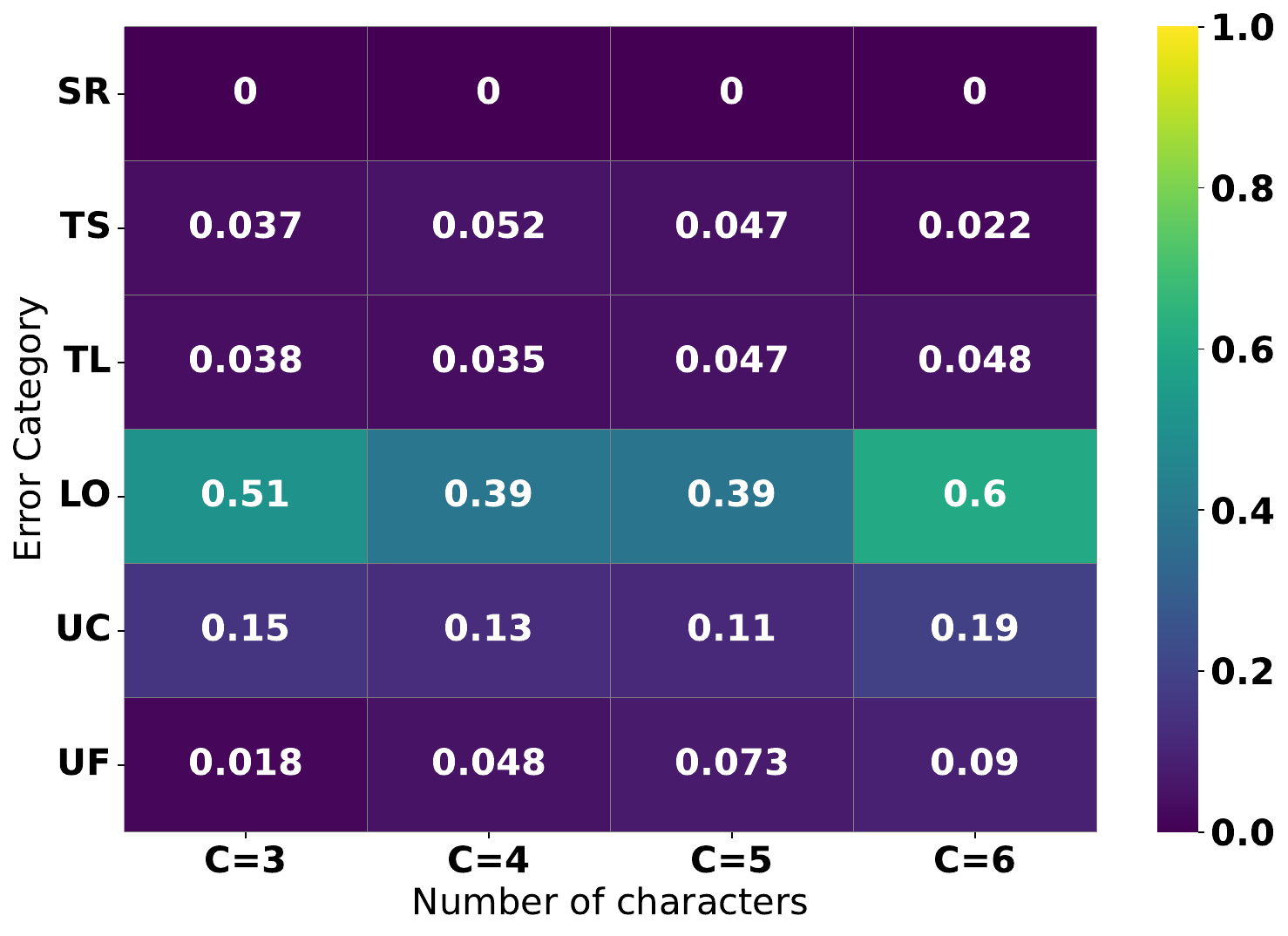}
        \caption{LLaMA-3-70B (four-shot CoT)}
        \label{fig:llama3_70b_four_cot_gpt4_reasoning_eval}
    \end{subfigure}
    \caption{Relative occurrences of the reasoning errors displayed by LLaMA-3-8B and LLaMA-3-70B when prompted via four-shot chain-of-thought. Values are obtained from GPT-4-based evaluations. Error categories are abbreviated for a more comprehensive overview: \textit{(SR) False statement reproduction}, \textit{(TS) Assuming statements to be true}, \textit{(TL) Misunderstanding the concept of truth and lies}, \textit{(LO) Misunderstanding logical operators}, \textit{(UC) Unjustified conclusion}, and \textit{(UF) Unfaithfulness}.}
    \label{fig:ai_error_analysis_small}
\end{figure}

\paragraph{Error Analysis}\label{subsec:error_analysis}
Figure \ref{fig:human_error_analysis} compares the relative occurrence of each error type, as outlined in Table \ref{tab:error_descriptions}, between LLaMA-3-8B (zero-shot) and LLaMA-3-70B (four-shot CoT). The values, derived from human annotations, are averaged across all statement sets for each number of characters. We find that LLaMA-3-8B exhibits a variety of errors, such as \emph{misunderstanding the concept of truth and lies (TL)} and \emph{unfaithfulness (UF)}. In contrast, LLaMA-3-70B predominantly struggles with deducing the logical implications of potentially false statements \emph{(LO)}. This trend—where lower-performing models show a wider array of errors, while higher-performing models predominantly struggle with logical deductions from statements that may be false—is further supported by our complementary analysis using GPT-4, as illustrated in Figure \ref{fig:ai_error_analysis_small}. We find that the error distribution obtained through GPT-4 positively correlates with the distribution obtained via manual labeling. Pearson correlation coefficients and their corresponding p-values are provided in Table \ref{tab:pearson_correlation} in the appendix. Additional results for LLaMA-2-7B and LLaMA-3-70B with zero-shot prompting are presented in Figure \ref{fig:ai_error_analysis} in the appendix. Further details of our automated error analysis can be found in Appendix \ref{appendix:results_ai_assisted_evaluation}.

\section{Conclusion}\label{sec:conclusion}
In this paper we introduce \emph{TruthQuest}, a benchmark for suppositional reasoning based on \emph{knights and knaves} puzzles. We demonstrate that LLMs exhibit significant difficulties solving these tasks. Our error analysis reveals that less proficient LLMs exhibit diverse errors, often failing to grasp the concept of truth and lies. In contrast, more proficient models primarily struggle with logical deductions from potentially false statements.

\section{Limitations}\label{sec:limitations}
While we introduce \emph{TruthQuest}, a novel benchmark designed to evaluate the suppositional reasoning capabilities of large language models, several limitations remain that could be addressed in future research.

\paragraph{Task Setup} Currently, \emph{TruthQuest} includes only \emph{knights and knaves} puzzles with a single, unique solution. Future work could expand this restriction to examine the impact of none or several solutions on model performance and behavior. Additionally, the benchmark is limited to simple propositional statements, as outlined in Table \ref{tab:statement_types}. Future iterations could incorporate more complex statement types that require more advanced inferences. Variations of \emph{knights and knaves} puzzles, which consider additional characters or altered character attributes, also present opportunities for further exploration. For instance, \citet{JOHNSONLAIRD199069} propose problems involving two types of persons: logicians, who always make valid deductions, and politicians, who never make valid deductions. An example problem states: ``A says that either B is telling the truth or else is a politician (but not both). B says that A is lying. C deduces that B is a politician. Is C a logician?'' Such variations represent compelling directions for future research.

\paragraph{Evaluation Framework} Our manual evaluation framework is constrained by the number of annotators and the volume of annotations provided. Despite our efforts to optimize available resources, these constraints may impact the scalability and generalizability of our results. While our automatic evaluation procedure offers a promising alternative, we found that error annotations obtained through this method exhibit only fair overall agreement with human annotations at the instance level (see Section \ref{appendix:ai_assisted_evaluation} in the appendix for further details). Additionally, although we consider various prompting techniques in our study, future research could explore the impact of more advanced methods, such as Tree-of-Thoughts~\cite{NEURIPS2023_271db992} or Graph-of-Thoughts~\cite{GoT}, on model performance.

\section*{Acknowledgments}
We express our gratitude to the members of the MaiNLP lab for their invaluable and insightful feedback. We specifically appreciate the suggestions of Diego Frassinelli, Michael Hedderich, Siyao Peng, Robert Litschko, Beiduo Chen, Jian Lan, Xinpeng Wang, Verena Blaschke, Elena Senger, and Rob van der Goot. Additionally, we would like to thank Huangyan Shan and Sebastian Loftus for their exceptional work and support in data annotation. We are also grateful to the anonymous reviewers for their constructive feedback. The emojis featured in Figure \ref{fig:knights_and_knaves_example_1} are designed by \href{https://openmoji.org/}{OpenMoji} – the open-source emoji and icon project (License: \href{https://creativecommons.org/licenses/by-sa/4.0/#}{CC BY-SA 4.0}). Lastly, we recognize the support for BP through the ERC Consolidator Grant
101043235.

\bibliography{anthology, custom}

\appendix

\section{Experimental Setup}\label{appendix:experimental_setup}
In this section, we provide additional details about the experimental setup. First, we elaborate on the language models employed in this study. Subsequently, we provide a detailed description of each error category devised to assess the models' reasoning.

\begin{table*}[tbhp]
\centering
\renewcommand{\arraystretch}{1.2}
{\fontsize{8}{10}\selectfont
\begin{tabular}
{>{\centering\arraybackslash}m{2.4cm} >{\centering\arraybackslash}m{1.4cm} >{\centering\arraybackslash}m{1.4cm} >{\centering\arraybackslash}m{1.5cm} >{\centering\arraybackslash}m{1.4cm} >{\centering\arraybackslash}m{1.4cm} >{\centering\arraybackslash}m{1.5cm} >{\centering\arraybackslash}m{1.4cm}}
\toprule
 Model & Base Model & Parameters & Context Length & Tokens & GPU hours & Carbon Emitted & Fine-tuning \\ 
 \midrule
LLaMA-2-7B-Chat & LLaMA-2 & 7B  & 4K tokens & 2.0T & 184K & 31 & SFT, RLHF \\
LLaMA-2-13B-Chat & LLaMA-2 & 13B  & 4K tokens & 2.0T & 369K & 62 & SFT, RLHF \\
LLaMA-2-70B-Chat & LLaMA-2 & 70B  & 4K tokens & 2.0T & 1.7M & 291 & SFT, RLHF \\
\cmidrule{1-8}
LLaMA-3-8B-Instruct & LLaMA-3 & 8B  & 8K tokens & 15T+ & 1.3M & 390 & SFT, RLHF \\
LLaMA-3-70B-Instruct & LLaMA-3 & 70B  & 8K tokens & 15T+ & 6.4M & 1900 & SFT, RLHF \\
\cmidrule{1-8}
Mixtral-8x7B-Instruct & Mixtral-8x7B & 46.7B & 32K tokens & - & - & - & SFT, DPO\\
\bottomrule
\end{tabular}
}
\caption{Details about the models used in this study. Tokens refer to the number of tokens in the pre-training data. Similarly, the context length, GPU hours and carbon emissions relate to the base model's pre-training. Carbon emissions are reported as tCO2eq. We use the following abbreviations for fine-tuning: supervised fine-tuning (SFT), reinforcement learning with human feedback (RLHF), direct preference optimization (DPO). Information about Llama 2 is taken from \citet{touvron2023llama}, while properties of Llama 3 are reported by \citet{meta2024llama3}. For Mixtral-8x7B, we consider the blog post of \citet{mistral2023mixtral}. Dashes denote unavailable information.}
\label{tab:llm_comparison}
\end{table*}

\subsection{Language Models}\label{appendix:language_models}
As outlined in Section \ref{sec:exp_setup}, six distinct large language models from three open-access model families are evaluated in this study. Detailed information, including the number of parameters and the context length for each model, is provided in Table \ref{tab:llm_comparison}. Each model is prompted with a system message that offers context about the task setup and specifies the required response format. Following this, a user prompt containing the task description is given. The complete prompt is depicted in Figure \ref{fig:task_prompt}. For few-shot setups, examples are presented in dialogue format, with the desired response indicated using the assistant's special tokens. Model responses are generated using nucleus sampling, utilizing the models' default values as specified on the Huggingface Platform (top-$p = 0.9$, temperature $T = 0.6$).\footnote{Please refer to: \href{https://huggingface.co/meta-llama}{huggingface.co/meta-llama}, and \href{https://huggingface.co/mistralai/Mixtral-8x7B-Instruct-v0.1}{https://huggingface.co/mistralai/Mixtral-8x7B-Instruct-v0.1}}. All few-shot prompts as well as the code for prompting and evaluating models on \emph{TruthQuest} are publicly accessible at \href{https://github.com/mainlp/TruthQuest}{https://github.com/mainlp/TruthQuest}. Additionally, all model responses, along with human annotations regarding potential errors (see Section \ref{sec:exp_setup}), are available at \href{https://huggingface.co/datasets/mainlp/TruthQuest-Human-Annotations}{https://huggingface.co/datasets/mainlp/TruthQuest-Human-Annotations}.

\subsection{Error Categorization}\label{appendix:error_categories}
To gain a deeper understanding of the models' reasoning behavior, we develop six different error categories that encompass common errors observed in the models' reasoning. These categories are established through a preliminary manual examination of the models' responses. Detailed descriptions of each error category are provided in Table \ref{tab:error_descriptions}. It is important to note that these error categories are not meant to be exhaustive. Instead, they are intended to offer practical insights into the models' frequent failure modes. For clarity, an example for each error category is provided below:

\paragraph{False statement reproduction (SR)} Smaller, less advanced models, such as LLaMA-2-7B, often fail to accurately reproduce statements from the problem description. For example, a model may incorrectly reproduce Pinky's statement ``Greeny is a knight and Bluey is a knave'' as ``Bluey states that both Pinky and Greeny are knaves.''

\paragraph{Assuming statements to be true (TS)} Models sometimes overlook the possibility that statements may be false and instead assume the truth of each statement. For example, a model might directly conclude from Pinky's statement ``Greeny is a knight and Bluey is a knave'' that Greeny is indeed a knight and Bluey a knave, without considering the potential falsity of the statement.

\paragraph{Misunderstanding the concept of truth and lies (TL)} Some models demonstrate a lack of understanding regarding truth and falsehood. For instance, a model may incorrectly assume that knights lie or knaves tell the truth. Additionally, models sometimes erroneously assume that knights only make statements about other knights and knaves only about other knaves.

\paragraph{Misunderstanding logical operators (LO)} When processing false statements, such as Pinky's assertion that ``Greeny is a knight and Bluey is a knave'' (when Pinky is actually a knave), models often fail to infer the logical implications of the lie. Specifically, the model may not deduce that the possibilities are: (i) both Greeny and Bluey are knaves, (ii) both are knights, or (iii) Greeny is a knave and Bluey is a knight.

\paragraph{Unjustified Conclusion (UC)} Models occasionally draw conclusions without providing a valid justification. For example, a model may erroneously assert that Greeny is a knave without offering any reasoning to support this conclusion.

\paragraph{Unfaithfulness (UF)} Models are sometimes inconsistent in their reasoning. For instance, a model might first deduce that Greeny is a knight, but later contradict this by asserting that Greeny is a knave, without addressing the discrepancy in its reasoning.

\begin{table*}[!ht]
\centering
{\fontsize{8}{10}\selectfont
\begin{tabular}{>{\centering\arraybackslash}p{0.1\textwidth}|>{\centering\arraybackslash}p{0.3\textwidth}|p{0.5\textwidth}}
\toprule
\textbf{Abbreviation} & \textbf{Error Category} & \multicolumn{1}{c}{\textbf{Description}} \\ 
\midrule
\cellcolor{red!15} \textbf{SR} & False statement reproduction & A problem statement is repeated incorrectly by the model.\\ 
\midrule
\cellcolor{cyan!15} \textbf{TS} & Assuming statements to be true & The possibility that statements might be lies is not considered.\\ 
\midrule
\cellcolor{yellow!15} \textbf{TL} & Misunderstanding the concept of truth and lies & Making false assumptions about the nature of truth-tellers or liars. For instance, the model mistakenly assumes that truth-tellers lie, while liars tell the truth.\\ 
\midrule
\cellcolor{green!15} \textbf{LO} & Misunderstanding logical operators & The logical implications of a potentially false statement are not properly deduced.\\ 
\midrule
\cellcolor{orange!15} \textbf{UC} & Unjustified Conclusion & A conclusion such as “X is a truth-teller/liar” is presented without proper justification.\\ 
\midrule
\cellcolor{blue!15} \textbf{UF} & Unfaithfulness & A new conclusion explicitly contradicts a conclusion previously drawn.\\ 
\bottomrule
\end{tabular}
}
\caption{Error categories and their respective descriptions.}
\label{tab:error_descriptions}
\end{table*}

\section{Additional Results}\label{appendix:additional_results}
We report additional results of the models' performance on \emph{TruthQuest} in Section \ref{appendix:task_performance}, and provide a supplementary analysis of the errors they commonly display in Section \ref{appendix:error_analysis}.

\subsection{Task Performance}\label{appendix:task_performance}
Table \ref{tab:full_accuracy_values} supplements Table \ref{tab:accuracies} by displaying LLaMA-3-8B's task performance for different prompting techniques. We observe that similar to LLaMA-3-70B, chain-of-thought prompting can yield notable performance gains for problems of lower complexity, i.e. \ fewer number of characters. Similarly, other prompting techniques such as few-shot prompting or zero-shot CoT do not seem to increase model performance.

\subsubsection{Content Effects}\label{appendix:content_effects}
\begin{figure*}[t!]
    \centering
    \begin{subfigure}[t]{0.32\textwidth}
        \centering
        \includegraphics[width=\linewidth]{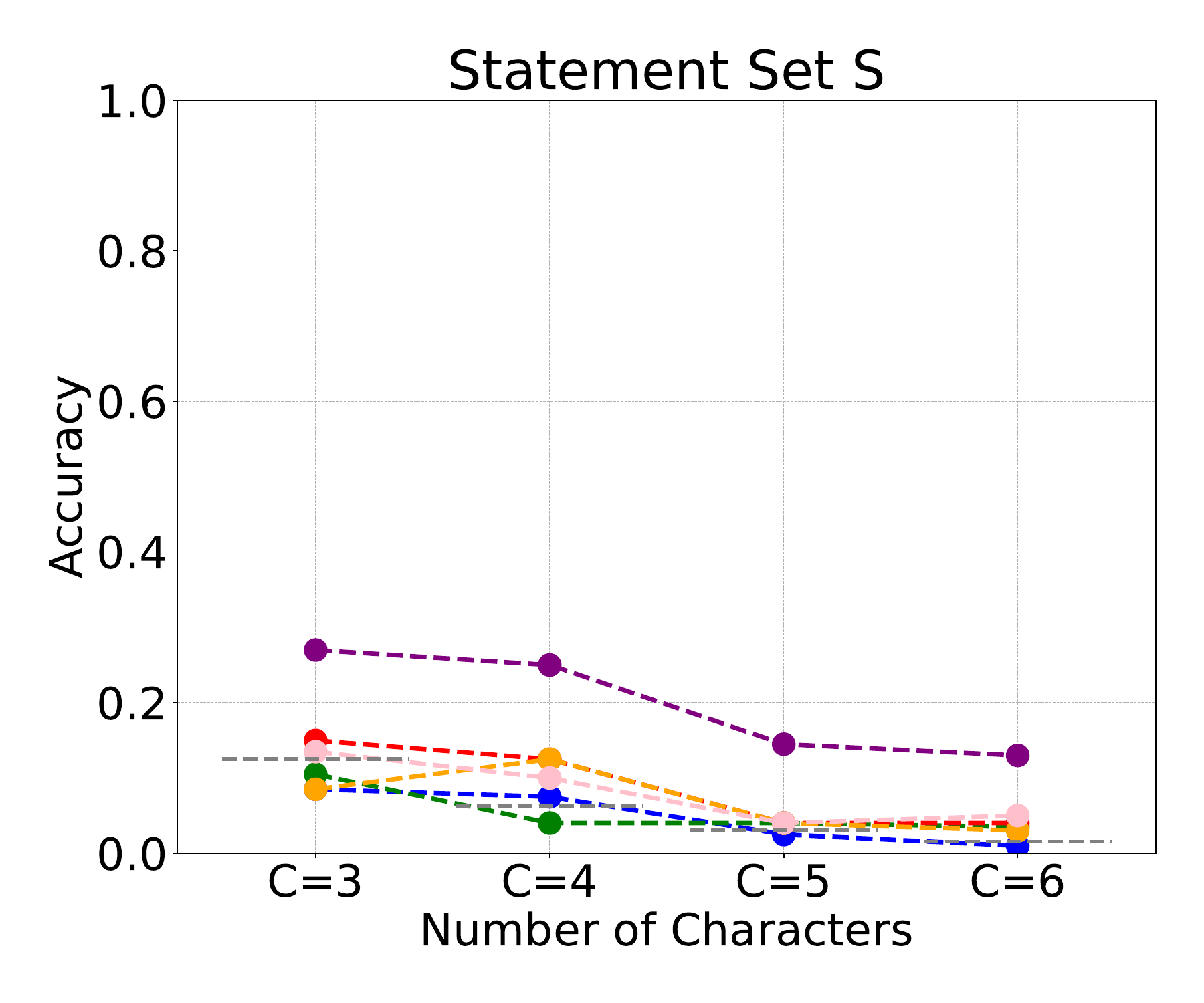}
        \caption{\emph{knights} and \emph{knaves}}
    \end{subfigure}\hfill
    \begin{subfigure}[t]{0.32\textwidth}
        \centering
        \includegraphics[width=\linewidth]{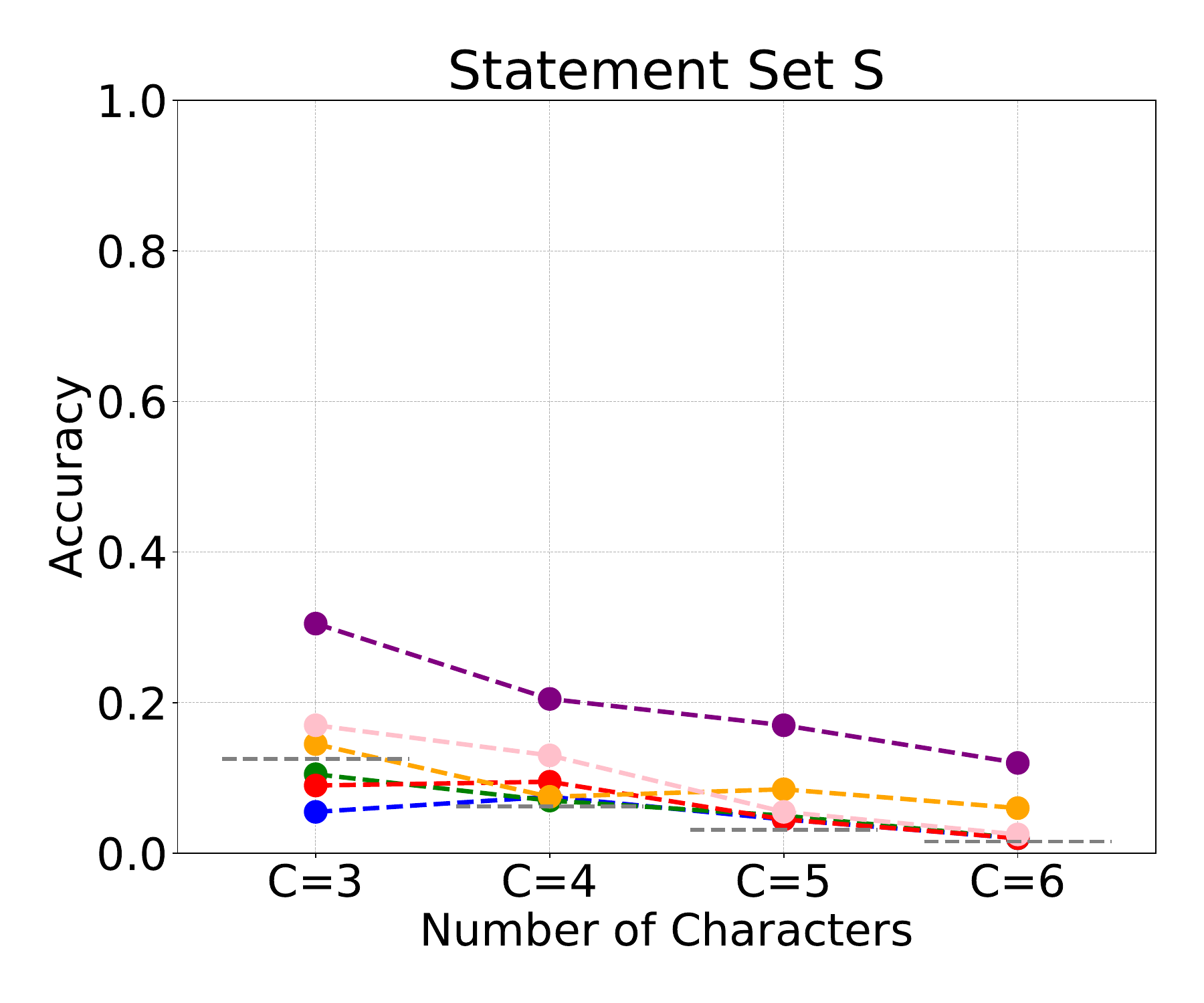}
        \caption{\emph{truth-tellers} and \emph{liars}}
    \end{subfigure}\hfill
    \begin{subfigure}[t]{0.32\textwidth}
        \centering
        \includegraphics[width=\linewidth]{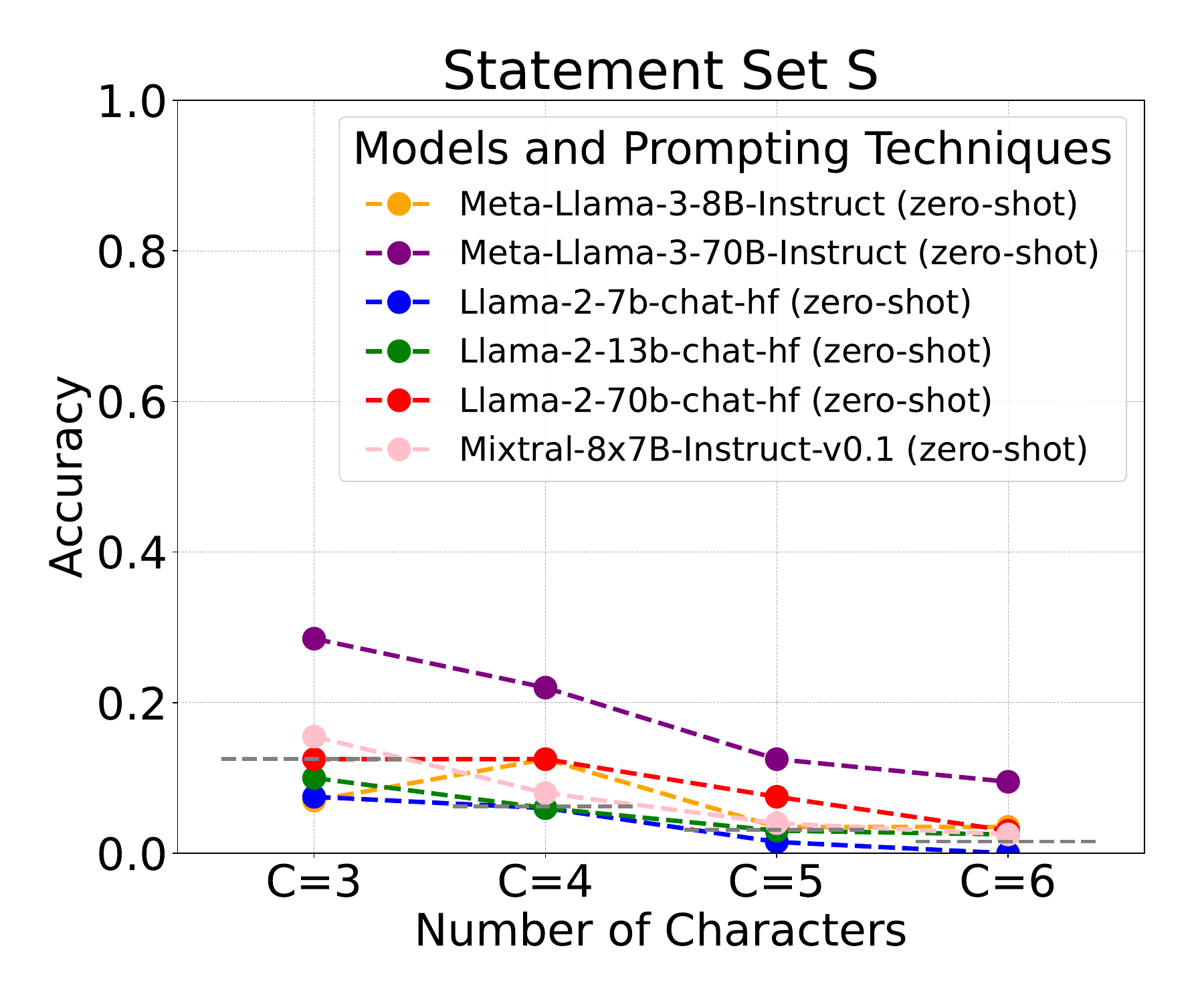}
        \caption{\emph{jabbas} and \emph{tettes}}
    \end{subfigure}\\[1em]
    \begin{subfigure}[t]{0.32\textwidth}
        \centering
        \includegraphics[width=\linewidth]{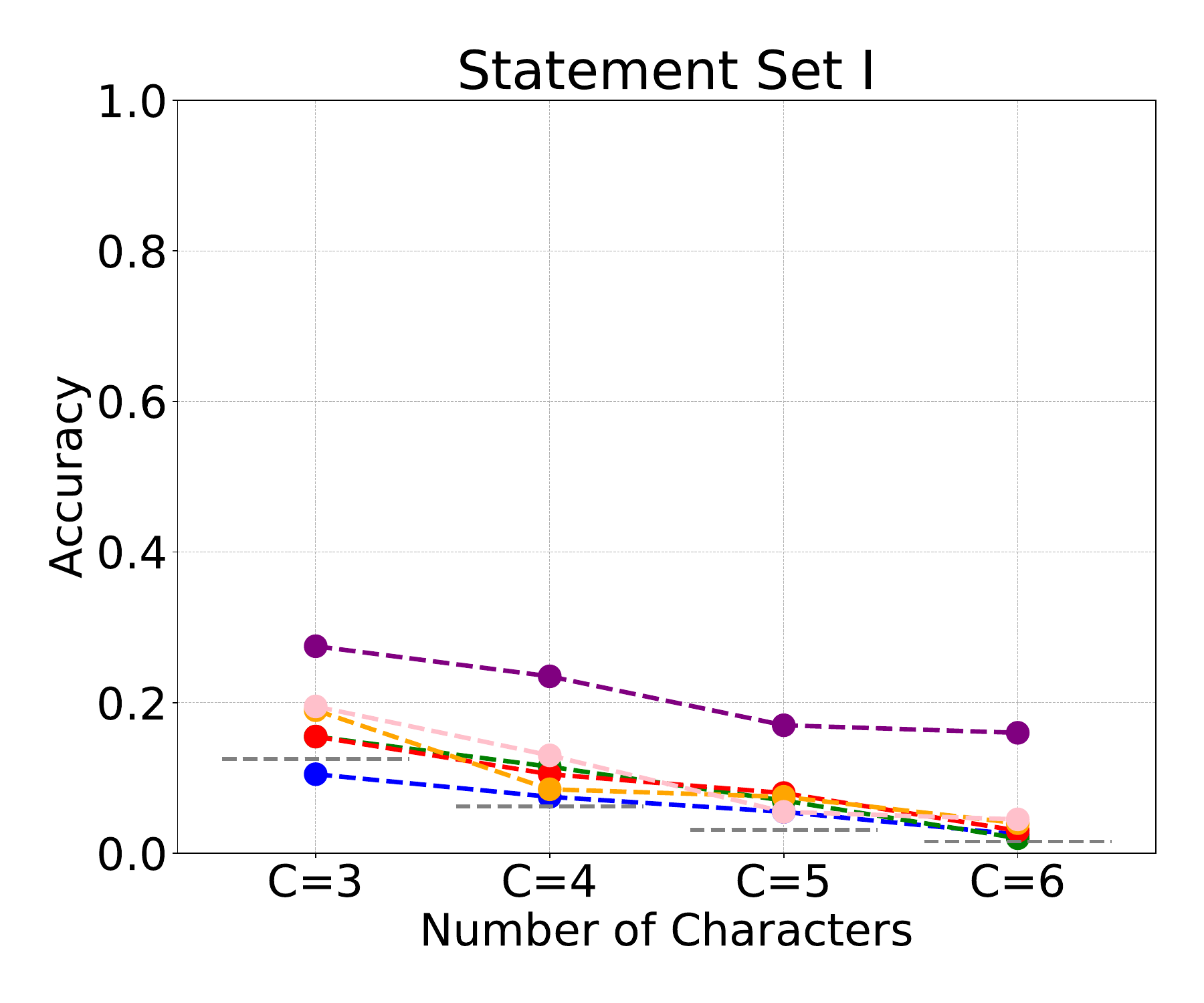}
        \caption{\emph{knights} and \emph{knaves}}
    \end{subfigure}\hfill
    \begin{subfigure}[t]{0.32\textwidth}
        \centering
        \includegraphics[width=\linewidth]{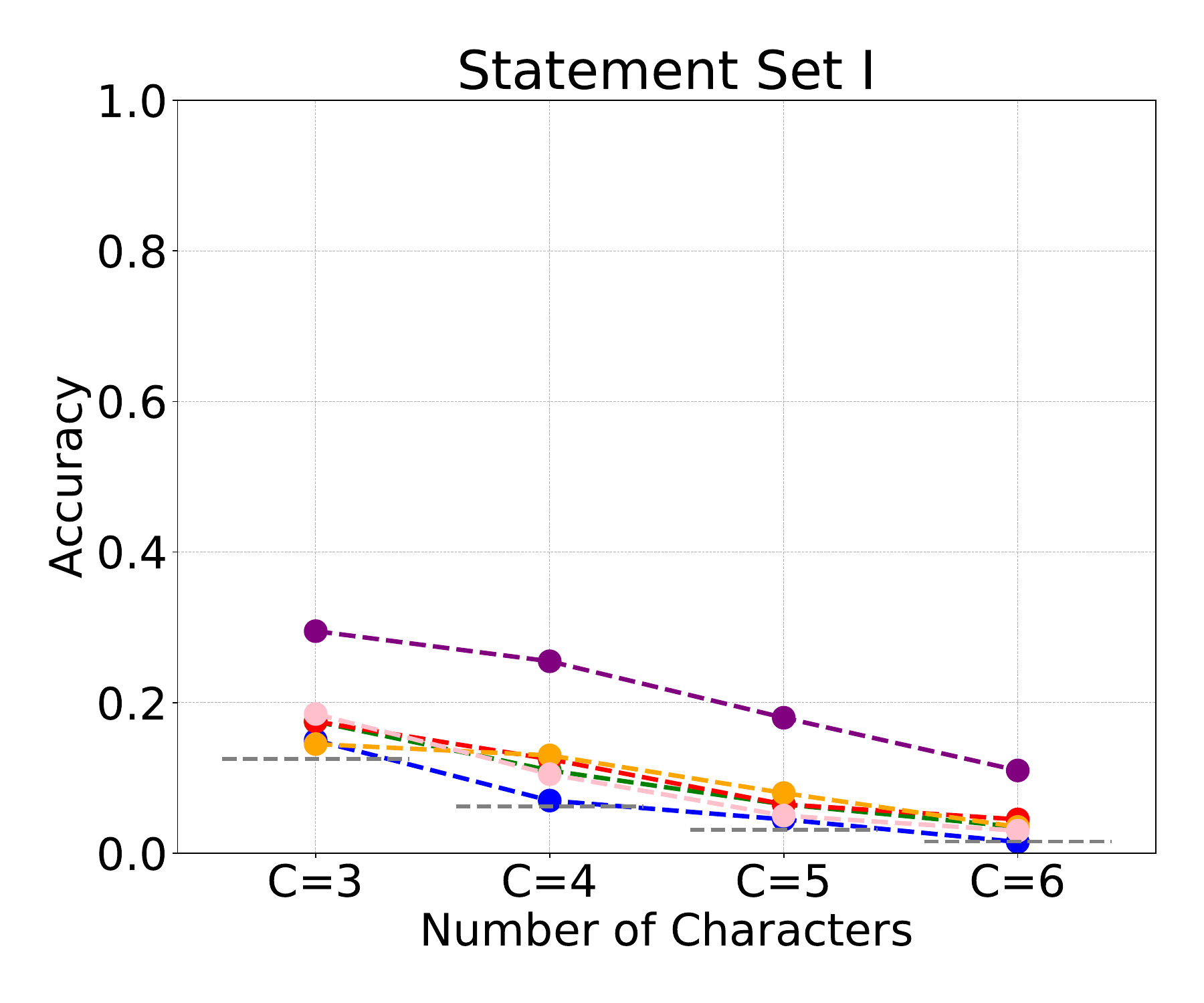}
        \caption{\emph{truth-tellers} and \emph{liars}}
    \end{subfigure}\hfill
    \begin{subfigure}[t]{0.32\textwidth}
        \centering
        \includegraphics[width=\linewidth]{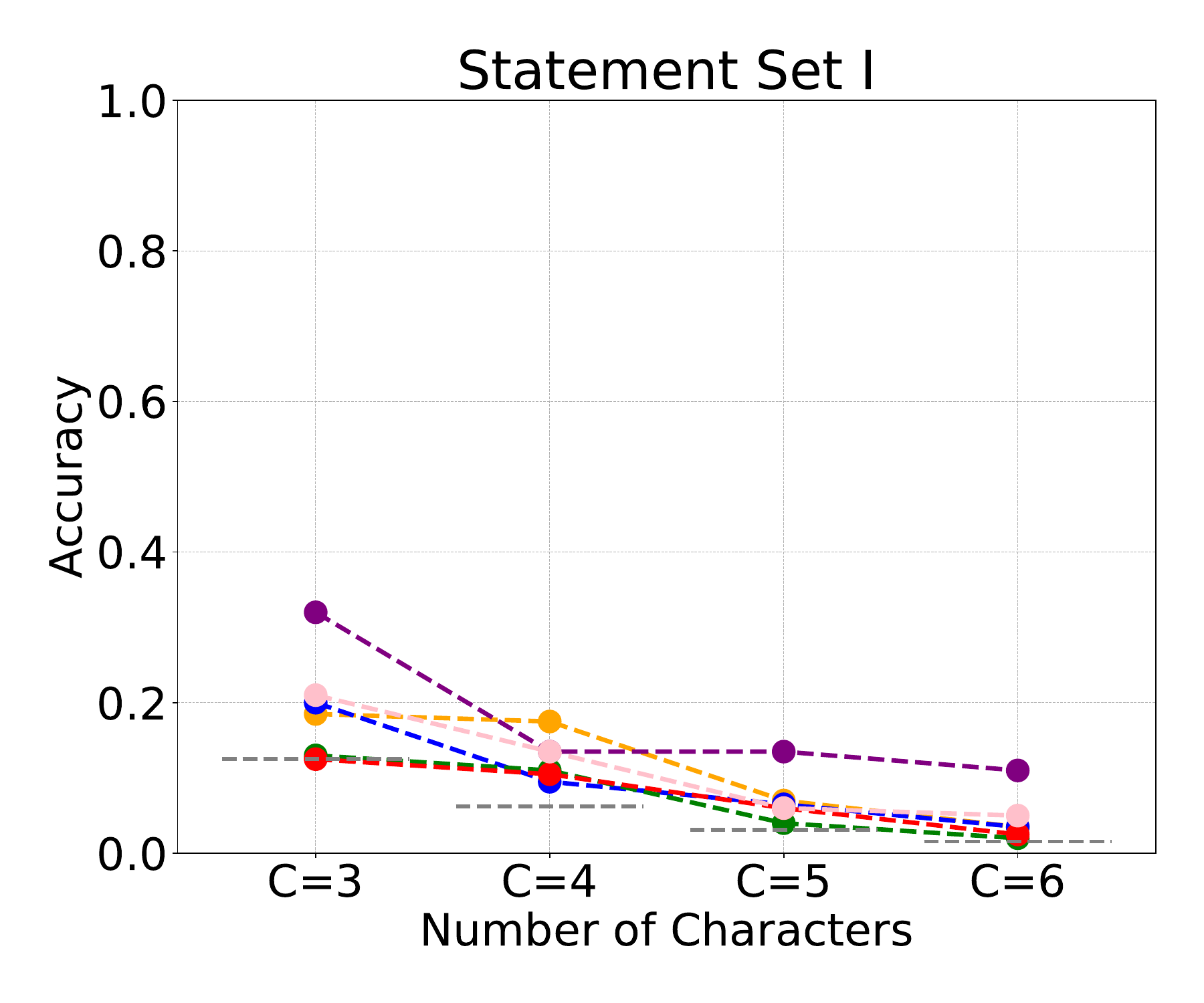}
        \caption{\emph{jabbas} and \emph{tettes}}
    \end{subfigure}\\[1em]
    \begin{subfigure}[t]{0.32\textwidth}
        \centering
        \includegraphics[width=\linewidth]{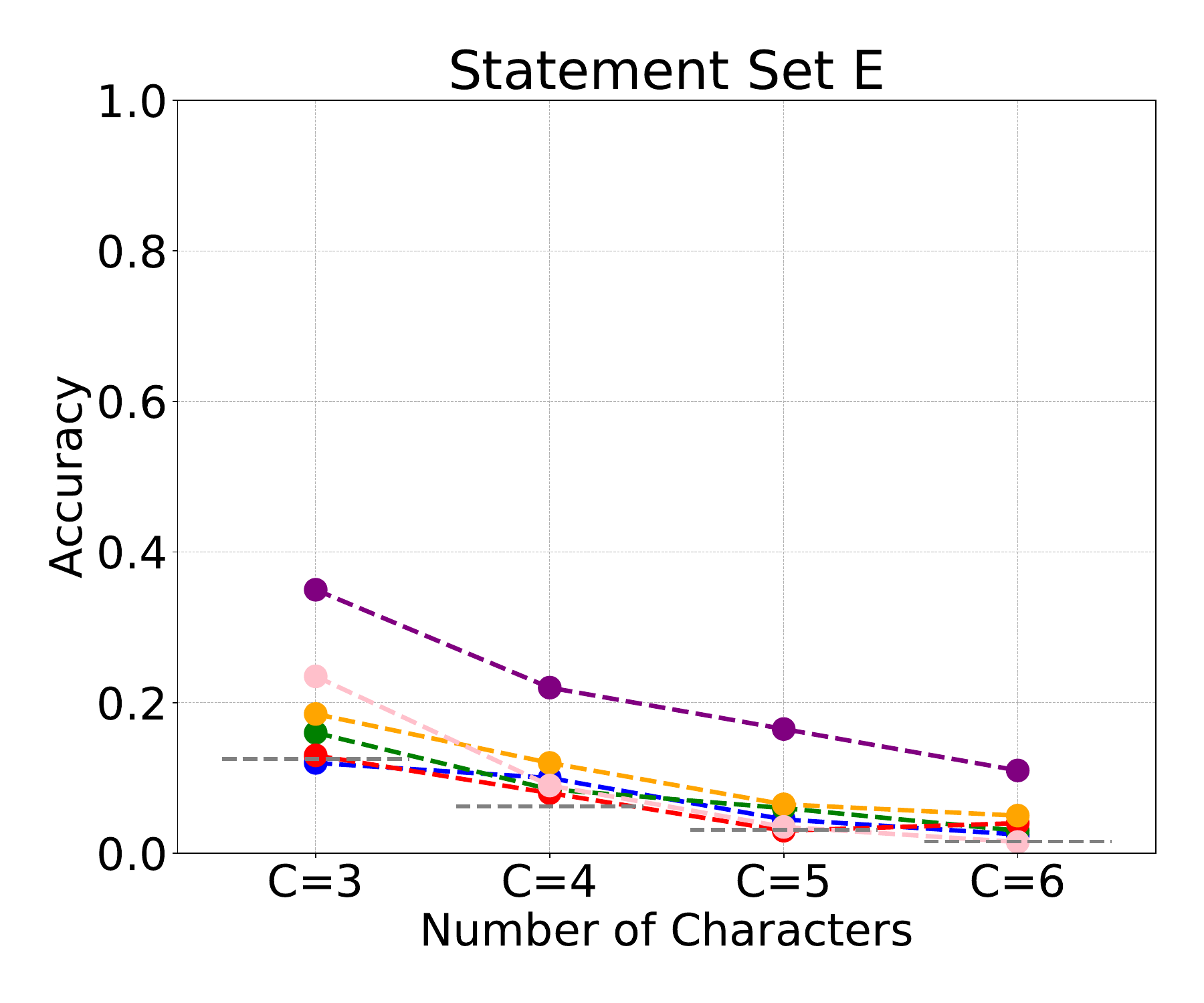}
        \caption{\emph{knights} and \emph{knaves}}
    \end{subfigure}\hfill
    \begin{subfigure}[t]{0.32\textwidth}
        \centering
        \includegraphics[width=\linewidth]{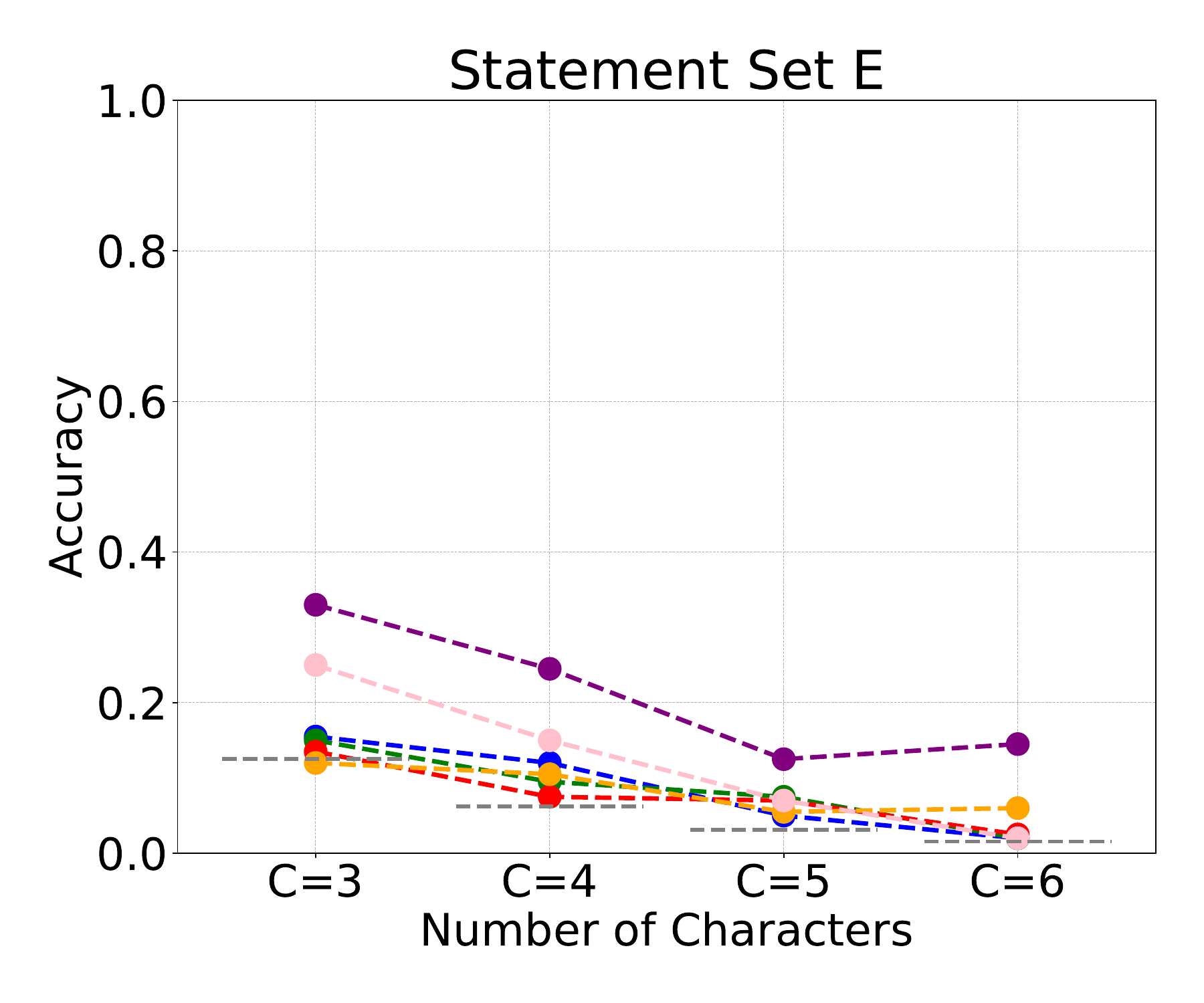}
        \caption{\emph{truth-tellers} and \emph{liars}}
    \end{subfigure}\hfill
    \begin{subfigure}[t]{0.32\textwidth}
        \centering
        \includegraphics[width=\linewidth]{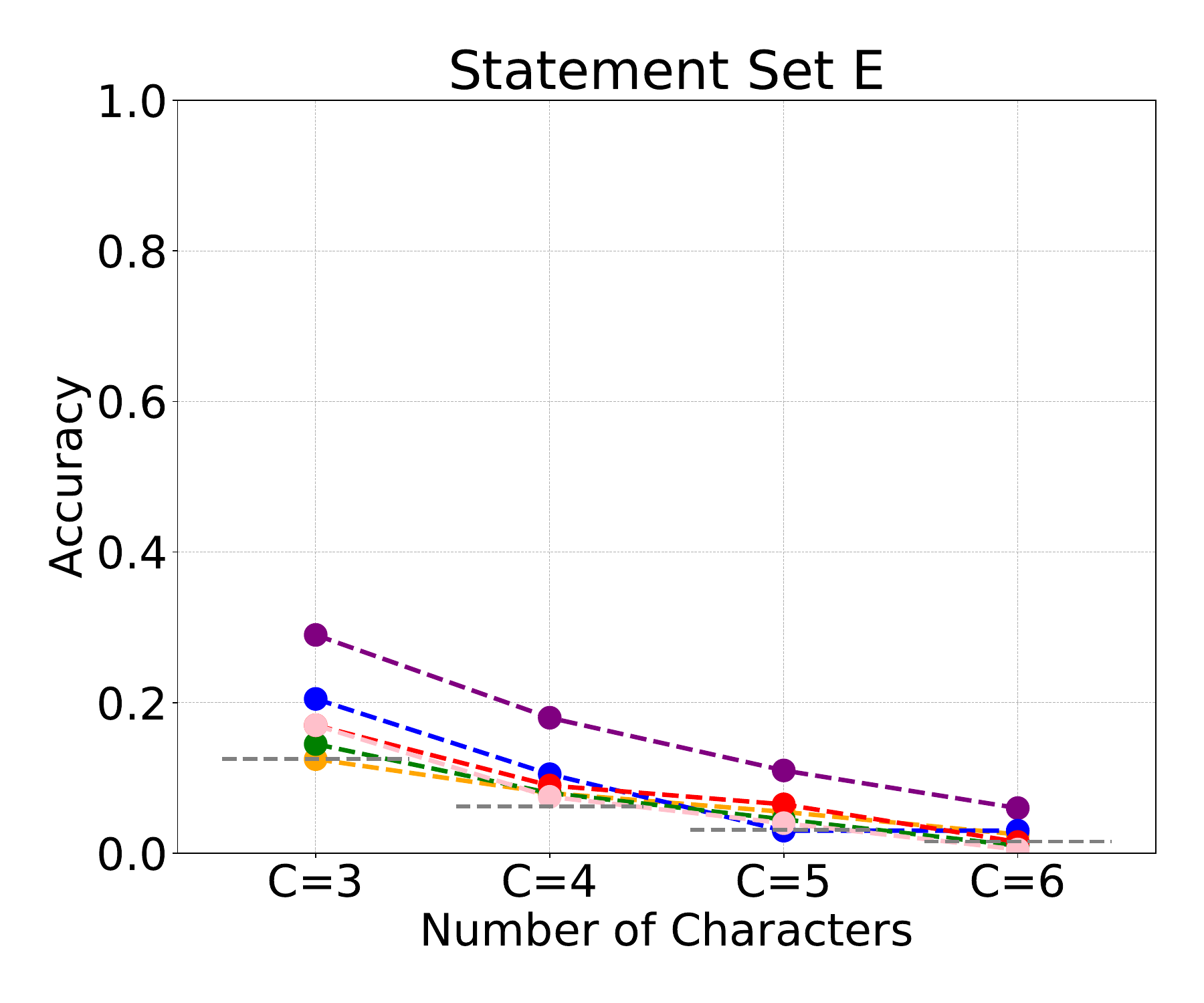}
        \caption{\emph{jabbas} and \emph{tettes}}
    \end{subfigure}
    \caption{Zero-shot performance of all models across the data subsets of \emph{TruthQuest}, focusing on the terminology used for characters who always tell the truth, and those who always lie. Three different formulations are compared: (i) the conventional \emph{knights} and \emph{knaves}, (ii) neutral descriptions such as \emph{truth-tellers} and \emph{liars}, and (iii) pseudo-terms such as \emph{jabbas} and \emph{tettes}.}
    \label{fig:content_effects}
\end{figure*}

In conventional \emph{knights and knaves} puzzles, \emph{knights} are characters who always tell the truth, while \emph{knaves} always lie. However, this setup can be modified by assigning different terms to these characters. It is likely that the models evaluated in this study have encountered conventional \emph{knights and knaves} puzzles during their training procedure, as such examples are readily available on the internet~\cite{smullyan1978whatis}. By altering the terms used for truth-tellers and liars, we can significantly reduce the likelihood that the models have been exposed to similar samples in our benchmark. Consequently, we analyze the impact of the terminology used for \emph{knights} and \emph{knaves} on model performance. Specifically, we examine three different formulations: (i) the conventional \emph{knights} and \emph{knaves}, (ii) neutral descriptions such as \emph{truth-tellers} and \emph{liars}, and (iii) pseudo-terms such as \emph{jabbas} and \emph{tettes}. Figure \ref{fig:content_effects} illustrates the zero-shot performance of all models for each terminology setup across the different subsets of \emph{TruthQuest}. Surprisingly, we find no substantial impact of the choice of terms on the models' task performance. We hypothesize that this may be because the specific instances generated for \emph{TruthQuest} have not yet been exposed to the internet. Consequently, the models might have encountered only a negligible fraction of instances by chance during their training process.

\subsection{Error Analysis}\label{appendix:error_analysis}
We examine the models' reasoning errors through both comprehensive manual inspections and AI-based evaluations of their rationales. The following sections present additional results from both evaluation methods.

\subsubsection{Human Evaluation}\label{appendix:results_human_evaluation}
\begin{figure}[b!]
  \centering 
  \includegraphics[width=0.48\textwidth]{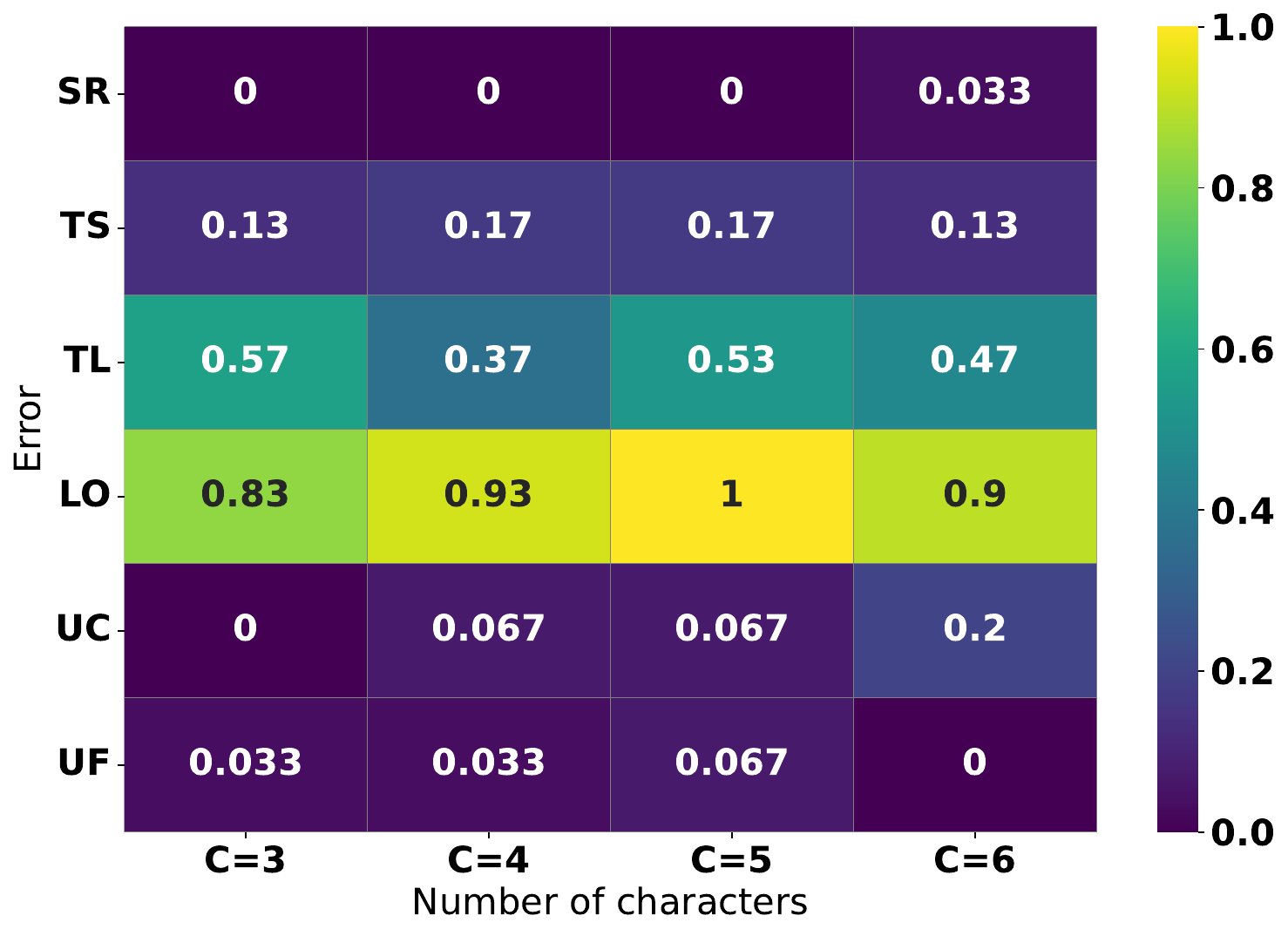}
  \caption{
  Relative occurrences of the reasoning errors displayed by LLaMA-3-70B when prompted in a zero-shot setting. Values are obtained from human evaluation. Error categories are abbreviated for a more comprehensive overview: \textit{(SR) False statement reproduction}, \textit{(TS) Assuming statements to be true}, \textit{(TL) Misunderstanding the concept of truth and lies}, \textit{(LO) Misunderstanding logical operators}, \textit{(UC) Unjustified conclusion}, and \textit{(UF) Unfaithfulness}.
  }
  \label{fig:human_error_analysis_llama_3_70B_zero_shot}
\end{figure}

As outlined in Section \ref{sec:exp_setup}, we manually evaluate a subset of the models' responses to asses their errors encountered during reasoning (for a detailed explanation of the evaluation procedure, please refer to Section \ref{appendix:human_evaluation}). To supplement our findings summarized in Figure \ref{fig:human_error_analysis}, , we report the common errors exhibited by LLaMA-3-70B when prompted in a zero-shot setting (see Figure \ref{fig:human_error_analysis_llama_3_70B_zero_shot}). We observe that, similar to LLaMA-3-8B (zero-shot) and LLaMA-3-70B (four-shot CoT), LLaMA-3-70B (zero-shot) frequently displays errors when deducing the logical implications of potentially false statements \emph{(LO)}. Additionally, we find that while the model struggles less with understanding the concept of truth and lies \emph{(TL)} compared to LLaMA-3-8B (zero-shot), it still exhibits this error category more frequently than LLaMA-3-70B (four-shot CoT). This trend, indicating that more proficient models better grasp the concept of truth and lies than lower-performing ones, is also reflected in our analysis of all model responses conducted via automatic LLM-based evaluation (for details, refer to Section \ref{appendix:results_ai_assisted_evaluation}).

\begin{figure*}[h!]
    \centering
    \begin{subfigure}[b]{0.47\textwidth}
        \centering
        \includegraphics[width=\linewidth]{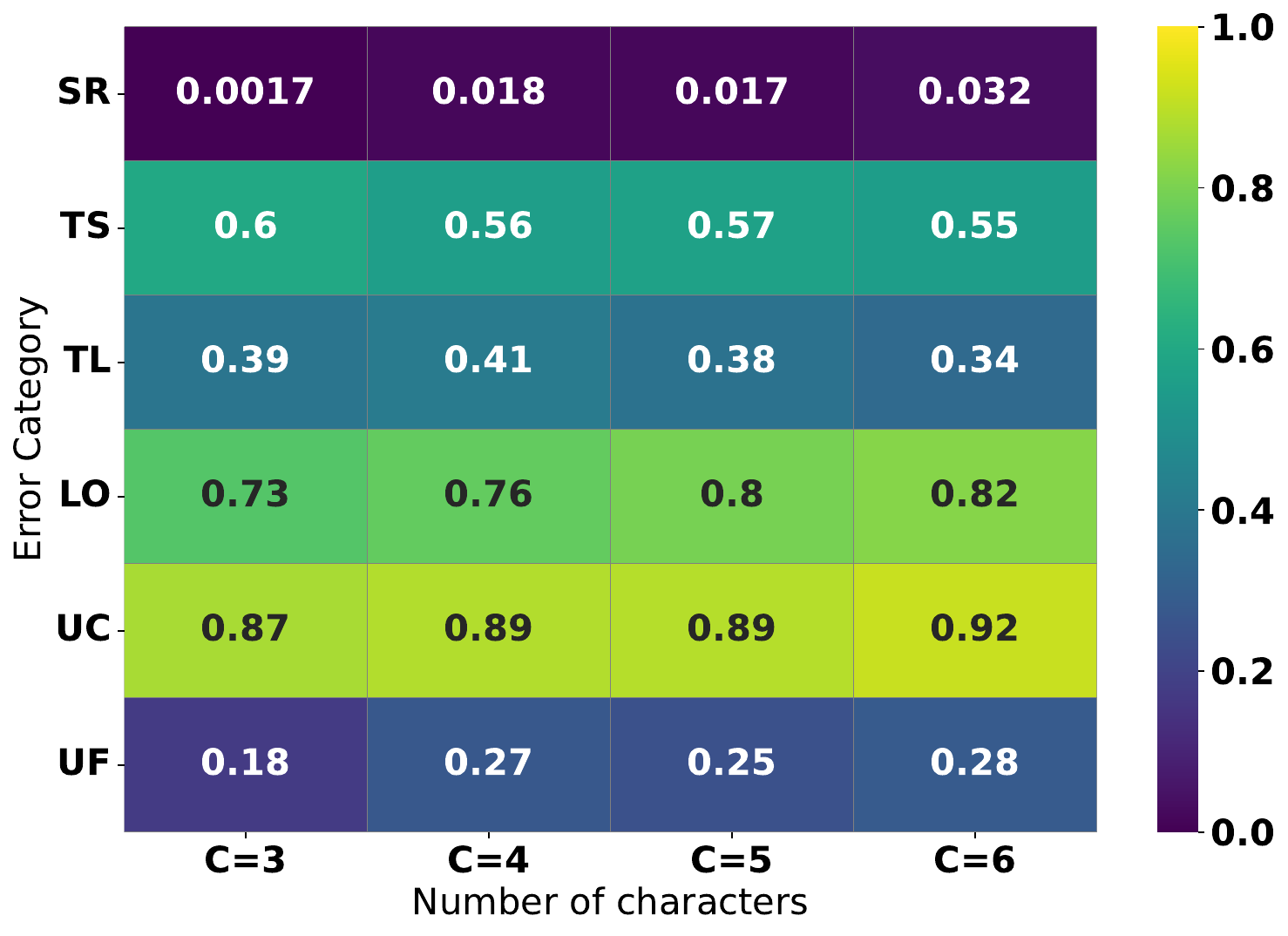}
        \caption{LLaMA-2-7B (zero-shot)}
    \end{subfigure}\hfill
    \begin{subfigure}[b]{0.47\textwidth}
        \centering
        \includegraphics[width=\linewidth]{imgs/results/ai_error_analysis/ai_overall_meta_llama_Meta_Llama_3_8B_Instruct_zero_shot.pdf}
        \caption{LLaMA-3-8B (zero-shot)}
    \end{subfigure}\\[1em]
    \begin{subfigure}[b]{0.47\textwidth}
        \centering
        \includegraphics[width=\linewidth]{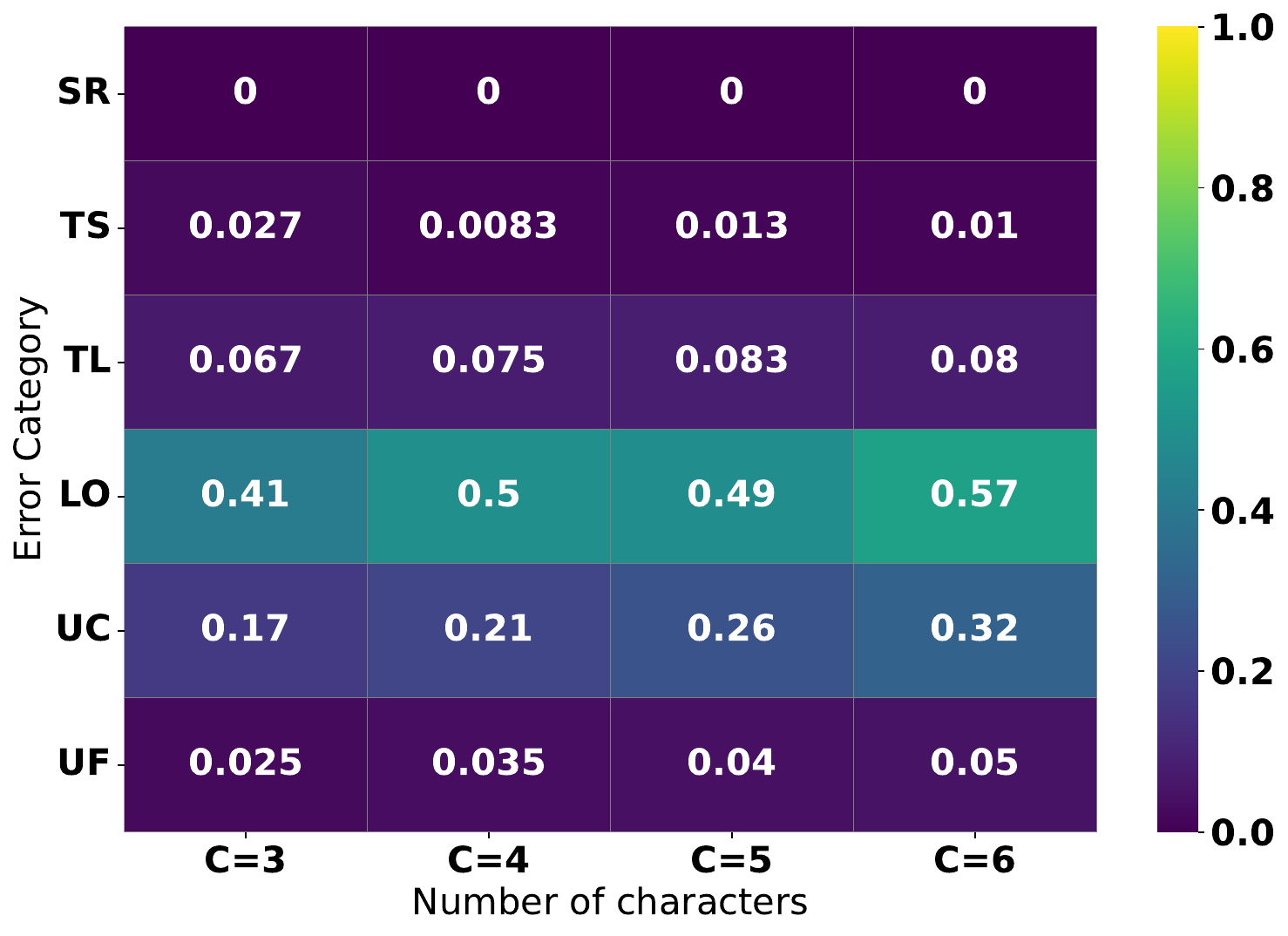}
        \caption{LLaMA-3-70B (zero-shot)}
    \end{subfigure}\hfill
    \begin{subfigure}[b]{0.47\textwidth}
        \centering
        \includegraphics[width=\linewidth]{imgs/results/ai_error_analysis/ai_overall_meta_llama_Meta_Llama_3_70B_Instruct_four_cot.pdf}
        \caption{LLaMA-3-70B (four-shot CoT)}
    \end{subfigure}
    \caption{Relative occurrences of the reasoning errors displayed by LLaMA-2-7B, LLaMA-3-8B, and LLaMA-3-70B in zero-shot prompting, as well as LLaMA-3-70B when prompted via four-shot chain-of-thought. Values are obtained from GPT-4-based evaluations. Error categories are abbreviated for a more comprehensive overview: \textit{(SR) False statement reproduction}, \textit{(TS) Assuming statements to be true}, \textit{(TL) Misunderstanding the concept of truth and lies}, \textit{(LO) Misunderstanding logical operators}, \textit{(UC) Unjustified conclusion}, and \textit{(UF) Unfaithfulness}.}
    \label{fig:ai_error_analysis}
\end{figure*}

\begin{table*}[tbp]
\centering
\renewcommand{\arraystretch}{1.2}
{\fontsize{8}{10}\selectfont
\begin{tabular}
{|>{\centering\arraybackslash}m{2.8cm}| *{7}{>{\centering\arraybackslash}m{1.4cm}|}}
\hline
  & \cellcolor{red!15} SR & \cellcolor{cyan!15} \textbf{TS}  & \cellcolor{yellow!15} \textbf{TL} & \cellcolor{green!15} \textbf{LO} & \cellcolor{orange!15} \textbf{UC} & \cellcolor{blue!15} \textbf{UF} \\ \hline
\hline
LLaMA-3-8B zero shot & $0.49$ & $0.33$ & $0.40$ & $0.68$ & $0.72$ & $0.53$  \\
\hline
LLaMA-3-70B zero shot & $-0.01$ & $0.37$ & $0.41$ & $0.51$ & $0.22$ & $0.15$  \\
\hline
LLaMA-3-70B four CoT & - & $0.90$ & $0.51$ & $0.74$ & $-0.04$ & $0.34$ \\
\hline
\end{tabular}
}
\caption{Cohen's Kappa values to assess the human inter-annotator agreement across different models, prompt setups and and error categories. Error categories are abbreviated for a more comprehensive overview: \textit{(SR) False statement reproduction}, \textit{(TS) Assuming statements to be true}, \textit{(TL) Misunderstanding the concept of truth and lies}, \textit{(LO) Misunderstanding logical operators}, \textit{(UC) Unjustified conclusion}, and \textit{(UF) Unfaithfulness}.}
\label{tab:human_cohens_kappa}
\end{table*}

\subsubsection{AI-Assisted Evaluation}\label{appendix:results_ai_assisted_evaluation}
High-quality human annotations are typically costly to obtain. In our study, we manually inspect 360 model responses from three different LLMs, where each instance is evaluated twice independently by two annotators (for details on the evaluation procedure, please refer to Section \ref{appendix:human_evaluation}). However, as our benchmark comprises 2,400 different instances, this evaluation procedure only covers a small subset of the models' responses. To complement our manual evaluation, we employ GPT-4 \cite{openai2024gpt4}\footnote{Specifically, version \texttt{gpt-4o-2024-05-13}.} to assess all 2,400 responses of a model with respect to the reasoning errors outlined in Table \ref{tab:error_descriptions} (for details about the exact prompts we employ, or the alignment between human and AI-based evaluation, please refer to Section \ref{appendix:ai_assisted_evaluation}). The respective results are illustrated in Figure \ref{fig:ai_error_analysis}. We present the relative occurrences of each error category for LLaMA-2-7B (zero-shot), LLaMA-3-8B (zero-shot), LLaMA-3-70B (zero-shot), and LLaMA-3-70B (four-shot CoT). All values are averaged across the different statement sets for each number of characters. Consistent with the results obtained through human evaluation, we observe a strong trend for higher-performing models to converge on errors related to deducing the correct logical implications of statements \emph{(LO)}. In contrast, lower-performing models such as LLaMA-2-7B (zero-shot) or LLaMA-3-8B (zero-shot), display diverse errors, ranging from misconceptions about truth and lies \emph{(TL)} to unjustified conclusions \emph{(UC)}. Notably, LLaMA-2-7B is the only model that frequently fails to consider statements as lies \emph{(TS)}.

\section{Evaluation Procedures}\label{appendix:evaluation}
In this study, we utilize two types of evaluation methods: human evaluations and AI-assisted evaluations. Below, we provide further details on each method, including the instructions given to human annotators and the process by which large language models are employed to generate similar annotations automatically. Finally, we assess the quality of our automatic evaluation procedures by comparing the results to the results obtained via manual assessment.

\subsection{Human Evaluation}\label{appendix:human_evaluation}
As outlined in Section \ref{sec:exp_setup}, we manually inspect 360 responses from LLaMA-3-8B (zero-shot) and LLaMA-3-70B (zero-shot and four-shot CoT). This manual evaluation is independently conducted by two hired students with expertise in data annotation. Both student annotators are compensated according to national standards.

\subsubsection{Annotator Instructions}\label{appendix:human_annotator_instructions}
To ensure high-quality annotations, we provide extensive training to both annotators. This training involves multiple sessions in which we introduce the annotators to \emph{knights and knaves} puzzles, asking them to solve these puzzles by hand to familiarize themselves with the task structure. Once the annotators are confident in solving puzzles of this style, we present exemplary responses from the models evaluated in this study. Together, we discuss notable behaviors and errors exhibited by the models. Next, we introduce the annotators to the six error categories outlined in Table \ref{tab:error_descriptions}. We proceed only when both annotators confirm their full understanding of each error type and have no further questions. The annotators are then tasked with independently annotating model responses. For each response, they parse the model’s conclusion and assign a binary label (yes/no) to each error category, indicating its presence or absence in the model's reasoning. Initially, the annotators work with practice examples to highlight and address any ambiguities in the annotation process. They only move on to labeling the actual model responses when they are confident in their understanding of the labeling process. To maintain high annotation quality, we ask both annotators to review their annotations, ensuring any potential errors in their annotations are accounted for.

\subsubsection{Inter-Annotator Agreement}
To assess the quality of our manual annotations, we calculate the inter-annotator agreement, reporting an overall Cohen's kappa value of $\kappa = 0.70$, which indicates substantial agreement between the two annotators. Table \ref{tab:human_cohens_kappa} presents Cohen's kappa values for each model and error type. We observe that the agreement rate varies across different categories, ranging from none to perfect agreement. Notably, the values for \emph{False statement reproduction (FS)} in LLaMA-3-70B (zero-shot) and \emph{Unjustified conclusion (FS)} in LLaMA-3-70B (four-shot CoT) are almost zero. This is likely due to a strong bias in the label distribution towards \emph{no} labels, as these errors rarely occur in these models. All human annotations are publicly available at \href{https://huggingface.co/datasets/mainlp/TruthQuest-Human-Annotations}{https://huggingface.co/datasets/mainlp/TruthQuest-Human-Annotations}.

\begin{table*}[thp]
\centering
{\fontsize{8}{10}\selectfont
\begin{tabular}{cccccc}
\toprule
\textbf{Model} & \textbf{Characters\_3} & \textbf{Characters\_4} & \textbf{Characters\_5} & \textbf{Characters\_6} \\
\midrule
LLaMA-3-8B Zero Shot & $0.673 \; \left(0.143\right)$ & $0.754 \; \left(0.084\right)$ & $0.8211 \; \left(0.045\right)$ & $0.859 \; \left(0.029\right)$ \\
\midrule
LLaMA-3-70B Zero Shot & $0.734 \; \left(0.097\right)$ & $0.858 \; \left(0.0289\right)$ & $0.762 \; \left(0.078\right)$ & $0.819 \; \left(0.046\right)$ \\
\midrule
LLaMA-3-70B Four CoT & $0.877 \; \left(0.022\right)$ & $0.857 \; \left(0.0294\right)$ & $0.923 \; \left(0.009\right)$ & $0.962 \; \left(0.002\right)$ \\
\bottomrule
\end{tabular}
}
\caption{Pearson correlation for the distribution of reasoning errors computed between the human and AI-based error analyses. The Pearson correlation coefficients are computed for different numbers of characters and models. Respective p-values are reported in parentheses.}
\label{tab:pearson_correlation}
\end{table*}

\subsection{AI-Assisted Evaluation}\label{appendix:ai_assisted_evaluation}
In addition to the human evaluation, we employ GPT-4 to assess the models' reasoning errors. Similar to the human annotators, GPT-4 is tasked with assigning binary labels (yes/no) to each error category described in Table \ref{tab:error_descriptions}, indicating the presence or absence of the error type in the model's reasoning. Additionally, GPT-4 is required to provide a justification for each label assigned. To ensure GPT-4's comprehension of each error category, we provide detailed descriptions in the model input. The full prompt can be found in Figure \ref{fig:error_evaluator_prompt}. Furthermore, we present six few-shot examples illustrating the desired annotation behavior (see Figures \ref{fig:example_1_error_evaluator_prompt} to \ref{fig:example_6_error_evaluator_prompt}). To assess the quality of the annotations, we compute the Pearson correlation for the error distributions of LLaMA-3-8B (zero-shot), LLaMA-3-70B (zero-shot), and LLaMA-3-70B (four-shot CoT) between the automatically obtained labels and the human annotations. All correlation coefficients and their corresponding p-values are reported in Table \ref{tab:pearson_correlation}. Overall, we find that the error distribution obtained through GPT-4 strongly correlates with the error distribution obtained via manual labeling. However, on an instance level, we observe only fair agreement, with an overall Cohen's kappa value of $\kappa = 0.34$. For our automatic evaluation of 9,600 model responses (4 models × 2,400 responses), the cost was approximately \$250. All annotations obtained through GPT-4 are publicly available at \href{https://huggingface.co/datasets/mainlp/TruthQuest-AI-Annotations}{https://huggingface.co/datasets/mainlp/TruthQuest-AI-Annotations}.

\paragraph{Task performance.} To assess the quality of our performance evaluation procedure depicted in Figure \ref{fig:conclusion_evaluator}, we compute the proportion of instances where the final conclusions derived from our two-step method match those reported by manual assessment.  We find an alignment of 100\%.

\section{Prompts}\label{appendix:prompts}
We present all prompts used in this study. The task prompt is shown in Figure \ref{fig:task_prompt}. Figure \ref{fig:conclusion_evaluator_prompt} illustrates the prompt for the two-step conclusion evaluator. Additionally, the system prompt for GPT-4 is provided in Figure \ref{fig:error_evaluator_prompt}, along with the few-shot examples in Figures \ref{fig:example_1_error_evaluator_prompt} to \ref{fig:example_6_error_evaluator_prompt}. Few-shot examples are designed to demonstrate the desired assistant behavior (highlighted in orange) in response to the corresponding user request (highlighted in blue). All prompts are publicly accessible at \href{https://github.com/mainlp/TruthQuest}{https://github.com/mainlp/TruthQuest}.

\begin{table*}[t!]
{\fontsize{8}{10}\selectfont
\centering
\begin{tabular}{c|c|cccc|cccc|cccc}
\toprule
\multirow{2}{*}{\textbf{Model}} & \multirow{2}{*}{\textbf{Mode}} & \multicolumn{4}{c|}{\textbf{Set S}} & \multicolumn{4}{c|}{\textbf{Set I}} & \multicolumn{4}{c}{\textbf{Set E}} \\
\cmidrule(lr){3-6} \cmidrule(lr){7-10} \cmidrule(lr){11-14}
& & \textbf{3} & \textbf{4} & \textbf{5} & \textbf{6} & \textbf{3} & \textbf{4} & \textbf{5} & \textbf{6} & \textbf{3} & \textbf{4} & \textbf{5} & \textbf{6} \\
\midrule
\textbf{Random Baseline} & - & 0.13 & 0.06 & 0.03 & 0.02 & 0.13 & 0.06 & 0.03 & 0.02 & 0.13 & 0.06 & 0.03 & 0.02 \\
\midrule
\textbf{LLaMA-2-7b} & \multirow{6}{*}{\textbf{zero shot}} & 0.08 & 0.06 & 0.02 & 0.00 & 0.20 & 0.10 & 0.07 & 0.04 & 0.21 & 0.11 & 0.03 & 0.03 \\
\textbf{LLaMA-2-13b} & & 0.10 & 0.06 & 0.03 & 0.03 & 0.13 & 0.11 & 0.04 & 0.02 & 0.15 & 0.08 & 0.05 & 0.01 \\
\textbf{LLaMA-2-70b} & & 0.13 & 0.13 & 0.08 & 0.03 & 0.13 & 0.11 & 0.06 & 0.03 & 0.17 & 0.09 & 0.07 & 0.02 \\
\textbf{LLaMA-3-8B} & & 0.07 & 0.13 & 0.04 & 0.04 & 0.19 & \textbf{0.18} & 0.07 & 0.04 & 0.13 & 0.08 & 0.06 & 0.03 \\
\textbf{LLaMA-3-70B} & & \textbf{0.29} & \textbf{0.22} & \textbf{0.13} & \textbf{0.10} & \textbf{0.32} & 0.14 & \textbf{0.14} & \textbf{0.11} & \textbf{0.29} & \textbf{0.18} & \textbf{0.11} & \textbf{0.06} \\
\textbf{Mixtral-8x7B} & & 0.16 & 0.08 & 0.04 & 0.03 & 0.21 & 0.14 & 0.06 & 0.05 & 0.17 & 0.08 & 0.04 & 0.01 \\
\midrule
\multirow{5}{*}{\textbf{LLaMA-3-8B}} & \textbf{four shot} & 0.13 & \textbf{0.14} & 0.04 & \textbf{0.05} & 0.19 & \textbf{0.14} & 0.08 & \textbf{0.05} & 0.18 & 0.09 & \textbf{0.09} & 0.02 \\
& \textbf{eight shot} & 0.14 & 0.09 & 0.07 & 0.04 & 0.23 & \textbf{0.14} & 0.07 & 0.05 & 0.19 & 0.10 & 0.07 & \textbf{0.04} \\
& \textbf{zero CoT} & 0.12 & 0.06 & 0.05 & 0.03 & 0.16 & 0.07 & 0.06 & 0.04 & 0.20 & 0.13 & 0.07 & 0.03 \\
& \textbf{four CoT} & 0.13 & 0.12 & 0.06 & 0.04 & 0.12 & 0.12 & 0.08 & 0.04 & 0.21 & 0.09 & 0.05 & 0.03 \\
& \textbf{eight CoT} & \textbf{0.27} & 0.13 & \textbf{0.08} & 0.03 & \textbf{0.26} & 0.12 & \textbf{0.12} & 0.05 & \textbf{0.26} & \textbf{0.15} & 0.08 & 0.03 \\
\midrule
\multirow{5}{*}{\textbf{LLaMA-3-70B}} & \textbf{four shot} & 0.22 & 0.25 & 0.19 & 0.13 & 0.24 & 0.21 & 0.13 & 0.10 & 0.32 & 0.22 & 0.11 & 0.05 \\
& \textbf{eight shot} & 0.22 & 0.21 & 0.16 & 0.09 & 0.32 & 0.25 & 0.07 & 0.09 & 0.27 & 0.20 & 0.10 & 0.02 \\
& \textbf{zero CoT} & 0.23 & 0.17 & 0.14 & 0.12 & 0.28 & 0.17 & 0.15 & 0.09 & 0.29 & 0.17 & \textbf{0.12} & 0.08 \\
& \textbf{four CoT} & 0.46 & \textbf{0.31} & \textbf{0.21} & 0.16 & \textbf{0.33} & \textbf{0.27} & 0.11 & \textbf{0.15} & \textbf{0.40} & \textbf{0.25} & \textbf{0.12} & \textbf{0.10} \\
& \textbf{eight CoT} & \textbf{0.60} & 0.26 & \textbf{0.21} & \textbf{0.20} & \textbf{0.33} & 0.20 & \textbf{0.15} & 0.12 & 0.37 & 0.20 & \textbf{0.12} & \textbf{0.10} \\
\bottomrule
\end{tabular}
}
\caption{Additional accuracy values of LLaMA-3-8B for different prompting techniques across each subset of \emph{TruthQuest}. Bold values represent highest performance among a group. The random baseline indicates the accuracy achieved by guessing the identity of each character.}
\label{tab:full_accuracy_values}
\end{table*}

\begin{figure*}[b!]
  \centering 
  \input{tikz/appendix/conclusion_evalutor}
  \caption{
  A schematic overview of the conclusion evaluator.
  }
  \label{fig:conclusion_evaluator}
\end{figure*}

\begin{figure*}[tbp]
  \centering 
  \input{tikz/appendix/task_prompt}
  \caption{
  The task prompt. Placeholders such as ``<\emph{number of characters}>'' and ``<\emph{statements}>'' are replaced by the corresponding input of the problem description. Note that the arrangement and usage of special tokens may vary depending on the specific language model employed.
  }
  \label{fig:task_prompt}
\end{figure*}

\begin{figure*}[tbp]
  \centering 
  \input{tikz/appendix/conclusion_evaluator_prompt}
  \caption{
  The three-shot conclusion evaluator prompt. Examples are presented in chat form, where the task prompt is depicted in blue, and the desired answer is exemplified in orange.
  }
  \label{fig:conclusion_evaluator_prompt}
\end{figure*}

\begin{figure*}[tbp]
  \centering 
  \input{tikz/appendix/error_evaluator_prompt}
  \caption{
  The system message of the six-shot error evaluator prompt.
  }
  \label{fig:error_evaluator_prompt}
\end{figure*}

\begin{figure*}[tbp]
  \centering 
  \input{tikz/appendix/example_1_error_evaluator}
  \caption{
   The first few-shot example of the error evaluator six-shot prompt. Examples are presented in chat form, where the task prompt is depicted in blue, and the desired answer is exemplified in orange.
  }
  \label{fig:example_1_error_evaluator_prompt}
\end{figure*}

\begin{figure*}[tbp]
  \centering 
  \input{tikz/appendix/example_2_error_evaluator}
  \caption{
   The second few-shot example of the error evaluator six-shot prompt. Examples are presented in chat form, where the task prompt is depicted in blue, and the desired answer is exemplified in orange.
  }
  \label{fig:example_2_error_evaluator_prompt}
\end{figure*}

\begin{figure*}[tbp]
  \centering 
  \input{tikz/appendix/example_3_error_evaluator}
  \caption{
   The third few-shot example of the error evaluator six-shot prompt. Examples are presented in chat form, where the task prompt is depicted in blue, and the desired answer is exemplified in orange.
  }
  \label{fig:example_3_error_evaluator_prompt}
\end{figure*}

\begin{figure*}[tbp]
  \centering 
  \input{tikz/appendix/example_4_error_evaluator}
  \caption{
  The fourth few-shot example of the error evaluator six-shot prompt. Examples are presented in chat form, where the task prompt is depicted in blue, and the desired answer is exemplified in orange.
  }
  \label{fig:example_4_error_evaluator_prompt}
\end{figure*}

\begin{figure*}[tbp]
  \centering 
  \input{tikz/appendix/example_5_error_evaluator}
  \caption{
  The fifth few-shot example of the error evaluator six-shot prompt. Examples are presented in chat form, where the task prompt is depicted in blue, and the desired answer is exemplified in orange.
  }
  \label{fig:example_5_error_evaluator_prompt}
\end{figure*}

\begin{figure*}[tbp]
  \centering 
  \input{tikz/appendix/example_6_error_evaluator}
  \caption{
  The sixth few-shot example of the error evaluator six-shot prompt. Examples are presented in chat form, where the task prompt is depicted in blue, and the desired answer is exemplified in orange.
  }
  \label{fig:example_6_error_evaluator_prompt}
\end{figure*}

\end{document}

%% file: tikz/introduction/intro_problem.tex
\begin{tikzpicture}[
    font=\fontsize{8}{9.6}\selectfont,
    ]
    
    \node[] (start) {};
    \node[below=0.01\textwidth of start] (wizard) {\includegraphics[width=0.1\textwidth]{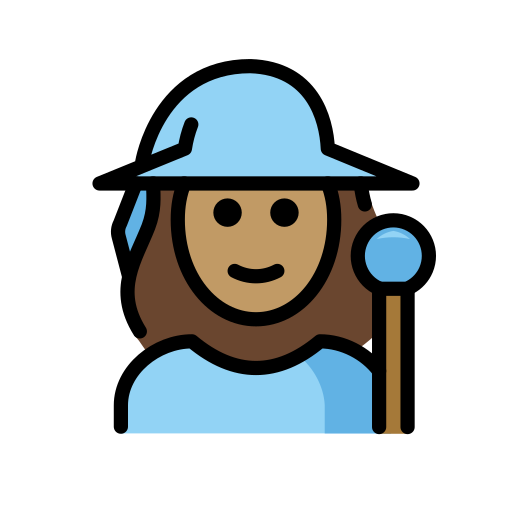}};
    \node[below=0.05\textwidth of wizard.south west, left=0.35\textwidth of wizard.south,  anchor=north] (person1) {\includegraphics[width=0.1\textwidth]{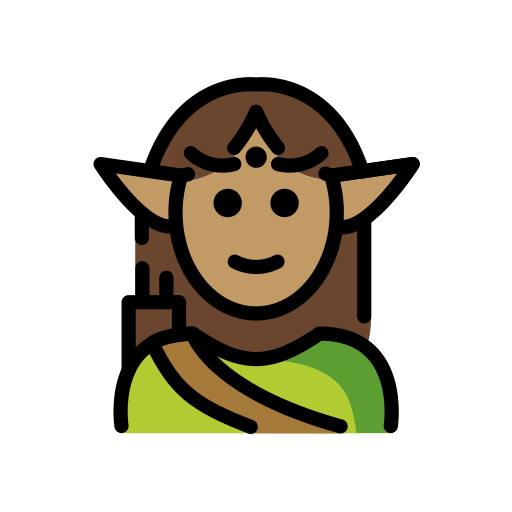}};
    \node[below=0.05\textwidth of person1.south, anchor=center] (person2) {\includegraphics[width=0.1\textwidth]{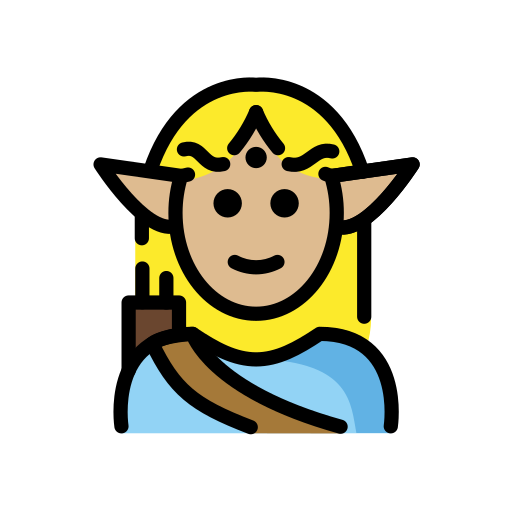}};
    \node[below=0.05\textwidth of person2.south, anchor=center] (person3) {\includegraphics[width=0.1\textwidth]{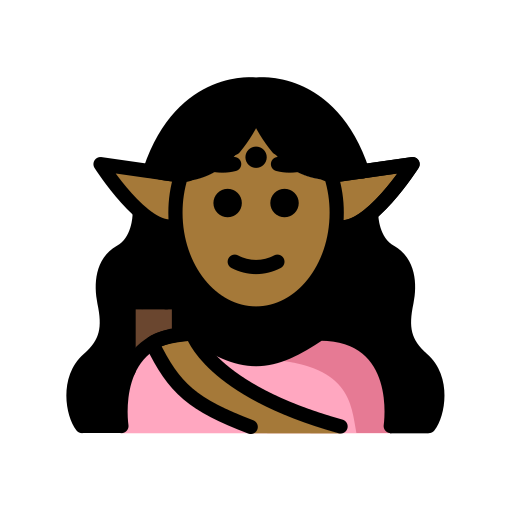}};

    \node[font=\fontsize{8}{9.6}\selectfont, below=-0.01\textwidth of person1.south, anchor=center] (text1) {Greeny};
    \node[font=\fontsize{8}{9.6}\selectfont, below=-0.01\textwidth of person2.south, anchor=center] (text2) {Bluey};
    \node[font=\fontsize{8}{9.6}\selectfont, below=-0.01\textwidth of person3.south, anchor=center] (text3) {Pinky};

    \node[rectangle callout, rounded corners, callout relative pointer={(0.3, 0.0)}, minimum width=0.2\textwidth, minimum height=1.5cm, left=0.05\textwidth of wizard, fill=cyan!20, text width=0.35\textwidth, align=left] at (wizard.center) {
    \textit{A special island is inhabited only by knights and knaves. Knights always tell the truth, while knaves always lie.\\
    \vspace{1mm}
    Based on the statements from three inhabitants, infer who is a knight and who is a knave.}
    };
    \node[rectangle callout, rounded corners, callout relative pointer={(-0.4,0.0)}, minimum width=3cm, minimum height=1.0cm, right=0.05\textwidth of person1, fill=orange!20, text width=0.35\textwidth, align=center] at (person1.center) {\textit{Pinky is a knave!}};
    \node[rectangle callout, rounded corners, callout relative pointer={(-0.4,0.0)}, minimum width=3cm, minimum height=1.0cm, right=0.05\textwidth of person2, fill=orange!20, text width=0.35\textwidth, align=center] at (person2.center) {\textit{I am a knight!}};
    \node[rectangle callout, rounded corners, callout relative pointer={(-0.4,0.0)}, minimum width=3cm, minimum height=1.0cm, right=0.05\textwidth of person3, fill=orange!20, text width=0.35\textwidth, align=center] at (person3.center) {\textit{Greeny is a knight and Bluey is a knave!}};
\end{tikzpicture}

%% file: tikz/appendix/conclusion_evalutor.tex
\begin{tikzpicture}
    font=\fontsize{8}{9.6}\selectfont,
    ]
    
    \tikzstyle{box} = [rectangle, rounded corners, minimum width=2cm, minimum height=1cm, text centered, draw=black, fill=cyan!20]
    \tikzstyle{arrow} = [thick,->,>=stealth]
    
    \node (model_response) [box, fill=orange!20, align=center] {Model \\ Response};
    \node (regex_parsing) [box, right=0.02\textwidth of model_response, align=center] {Regex\\Parsing};
    \node (extracted_conclusion) [box, fill=green!20, above right=0.02\textwidth and 0.1\textwidth of regex_parsing, align=center] {Extracted\\Conclusions};
    \node (failed_extractions) [box, fill=red!20, below right=0.02\textwidth and 0.1\textwidth of regex_parsing, align=center] {Failed\\Extractions};
    \node (llm_extractor) [box, fill=blue!20, right=0.02\textwidth of failed_extractions, align=center] {LLM\\Extractor};
    \node (regex_parsing_2) [box, fill=cyan!20, right=0.02\textwidth of llm_extractor, align=center] {Regex\\Parsing};
    \node (extracted_conclusion_2) [box, fill=green!20, right=0.02\textwidth of regex_parsing_2, align=center] {Extracted\\Conclusions};
    
    \draw [arrow] (model_response) -- (regex_parsing);
    \draw [arrow] (regex_parsing) -- ++(2,0) |- (extracted_conclusion);
    \draw [arrow] (regex_parsing) -- ++(2,0) |- (failed_extractions);
    \draw [arrow] (failed_extractions) -- (llm_extractor);
    \draw [arrow] (llm_extractor) -- (regex_parsing_2);
    \draw [arrow] (regex_parsing_2) -- (extracted_conclusion_2);

\end{tikzpicture}

%% file: tikz/appendix/task_prompt.tex
\begin{tikzpicture}[
    font=\fontsize{6}{7.6}\selectfont,
    ]
    
    \node[rectangle, rounded corners, minimum width=0.9\textwidth, minimum height=1.5cm, fill=cyan!20, text width=0.85\textwidth, align=left] (system_prompt) {
    [INST] <<SYS>> \\
    Your task is to solve a logical reasoning problem. You are given set of statements from which you must logically deduce the identity of a set of characters.\\
    \vspace{2mm}
    You must infer the identity of each character. First, explain your reasoning. At the end of your answer, you must clearly state the identity of each character by following the format:\\
    \vspace{2mm}
    CONCLUSION:\\
    A: ...\\
    B: ...\\
    C: ...\\
    ...\\
    <</SYS>>\\
    \#\#\# Instruction \#\#\#\\
    Assume that there exist only two types of people: knights and knaves. Knights always tell the truth, while knaves always lie.\\
    You are given the statements from <number of characters> characters. Based on their statements, infer who is a knights and who is a knave.\\
    \vspace{2mm}
    Based on the following statements, infer who is a knight and who is a knave:\\
    <statements>\\
    \vspace{2mm}
    First, explain your reasoning. End your answer by clearly stating the identity of each character in the following format:\\
    \vspace{2mm}
    A: knight/knave\\
    B: knight/knave\\
    C: knight/knave\\
    ... [/INST]
    };
\end{tikzpicture}

%% file: tikz/appendix/conclusion_evaluator_prompt.tex
\begin{tikzpicture}[
    font=\fontsize{6}{7.6}\selectfont,
    ]
    
    \node[rectangle, rounded corners, minimum width=0.9\textwidth, minimum height=1.5cm, fill=cyan!20, text width=0.85\textwidth, align=left] (system_prompt) {
    [INST] <<SYS>> \\
    Your task is to extract information from a given piece of text. You are given a piece of text that reasons about the identity of a set of characters. You must extract the final answer from the given text by clearly stating the deduced identity of each character in the following format:\\
    A: ...\\
    B: ...\\
    C: ...\\
    ...\\
    You must strictly follow this format. Do not generate any additional text. You must only extract information that is present in the text provided. Do not add information that goes beyond the one contained in the text at hand.\\
    <</SYS>>\\
    \#\#\# Instruction \#\#\#\\
    You are given a piece of TEXT that reasons about the identity of a set of characters. You must extract the FINAL ANSWER from the TEXT by clearly stating the deduced identity of each character in the following format:\\
    \#\#\# FINAL ANSWER:\\
    A: ... \\
    B: ... \\
    C: ... \\
    ...\\
    You must strictly follow this format. Do not generate any additional text. You must only extract information that is present in the text provided. Do not add information that goes beyond the one contained in the text at hand. If "not a" is placed in front of the identity, flip the result of that character, e.g. "not a knight" becomes "knave", while "not a knave" becomes "knight". You should simplify results like "knight/knight" to "knight" and "knave/knave" to "knave". However, do not simplify "knight/knave", or "knave/knight".\\
    \#\#\# TEXT: \\
    Great! Let's apply logical reasoning to deduce the identity of each character.\\
    A: Based on statement A, we know that C is a knave and D is a knave. This means that C and D are both knaves.\\
    B: Statement B tells us that C is a knave and D is a knave. Since C and D are both knaves, we can conclude that B is a knight.\\
    C: Statement C tells us that D is a knave. Since D is already identified as a knave in statements A and B, we can conclude that C is a knight.\\
    D: Statement D tells us that B is a knight. Since B has already been identified as a knight in statement B, we can confirm that D is a knave.\\
    Therefore, the identities of each character are:\\
    A: not a knight\\
    B: knight/knave\\
    C: knight/knight\\
    D: knave [/INST]
    };
    \node[rectangle, rounded corners, minimum width=0.9\textwidth, minimum height=1.0cm, fill=orange!20, text width=0.85\textwidth, align=left, below=0.0\textwidth of system_prompt.south, anchor=north] (answer1) {
    <|assistant|> \#\#\# FINAL ANSWER: \\
    A: knave \\
    B: knight/knave \\
    C: knight\\
    D: knave </|assistant|>
    };
    \node[rectangle, rounded corners, minimum width=0.9\textwidth, minimum height=1.0cm, fill=cyan!20, text width=0.85\textwidth, align=left, below=0.0\textwidth of answer1.south, anchor=north] (example2) {
    [INST] \#\#\# TEXT: \\
    Sure! Let's start by analyzing the statements one by one.\\                                                                                                                               1. A: B is a knave.\\
    Since A says that B is a knave, we can deduce that B is a knave.\\
    2. B: If D is a knight, then C is a knave.\\
    This statement implies that if D is a knight, then C is a knave. However, since B is a knave, we can deduce that D is not a knight. Therefore, C must be a knight.\\
    3. C: B is a knight.\\
    Since C says that B is a knight, we can deduce that B is a knight.\\
    4. D: A is a knave and C is a knave.\\
    This statement is not helpful in determining the identity of D, as D is simply repeating what A and C have said.\\
    Now, let's summarize our conclusions:\\
    A: knave\\
    B: knave\\
    C: knight\\
    D: unable to determine\\
    However, based on the statements, we can deduce that B and C are knaves, and A is a knight. Therefore, the correct identities of the characters are:
    A: knight\\
    B: knave\\
    C: knave\\
    D: unable to determine [/INST]
    };
    \node[rectangle, rounded corners, minimum width=0.9\textwidth, minimum height=1.0cm, fill=orange!20, text width=0.85\textwidth, align=left, below=0.0\textwidth of example2.south, anchor=north] (answer2) {
    <|assistant|> \#\#\# FINAL ANSWER:\\
    A: knight\\
    B: knave\\
    C: knave\\
    D: unable to determine </|assistant|>
    };
    \node[rectangle, rounded corners, minimum width=0.9\textwidth, minimum height=1.0cm, fill=cyan!20, text width=0.85\textwidth, align=left, below=0.0\textwidth of answer2.south, anchor=north] (example3) {
    [INST] \#\#\# TEXT: \\
    Okay, I understand the problem. Here's my reasoning:\\
    From statement A, if B is a knave, then C is a knight. This means that if B is a knave, then C must be a knight.\\
    From statement B, if A is a knave, then C is a knight. This means that if A is a knave, then C must be a knight.\\
    Now, let's examine the third statement\\
    C: B is a knave. If B is a knave, then C is a knave. But we know that knaves always lie, so we can't trust what C says. Therefore, we can't determine the identity of C based on this statement alone.\\
    However, we can use the first two statements to deduce the identity of C. If A is a knave, then C is a knight, and if B is a knave, then C is a knight. Since we don't know the identity of A and B, we can't determine the identity of C.\\
    CONCLUSION:\\
    A: knave/knight (cannot be determined)\\
    B: knave/knight (cannot be determined)\\
    C: knave [/INST]
    };
    \node[rectangle, rounded corners, minimum width=0.9\textwidth, minimum height=1.0cm, fill=orange!20, text width=0.85\textwidth, align=left, below=0.0\textwidth of example3.south, anchor=north] (answer3) {
    <|assistant|> \#\#\# FINAL ANSWER:\\
    A: knave/knight\\
    B: knave/knight\\
    C: knave </|assistant|>
    };
    \node[rectangle, rounded corners, minimum width=0.9\textwidth, minimum height=0.05\textwidth, fill=cyan!20, text width=0.85\textwidth, align=left, below=0.0\textwidth of answer3.south, anchor=north] (example3) {
    <Actual task description>
    };
\end{tikzpicture}

%% file: tikz/appendix/error_evaluator_prompt.tex
\begin{tikzpicture}[
    font=\fontsize{6}{7.6}\selectfont,
    ]
    
    \node[rectangle, rounded corners, minimum width=0.9\textwidth, minimum height=1.5cm, fill=cyan!20, text width=0.85\textwidth, align=left] (system_prompt) {
    <|system|> \\
    Your task is to classify whether a set of errors are present in the response of a large language model. You are given the initial PROBLEM STATEMENT and the model's RESPONSE.\\
    You must consider the following error types:\\
    \vspace{2mm}
    - False statement reproduction (A statement of the problem description is repeated incorrectly.)\\
    - Assuming statements to be true (The possibility that a character's statement is a lie is not considered. The model does not question the premises' truth but assumes it as given.)\\
    - Misunderstanding the concept of truth and lies (The model does not understand that a knights always tells the truth and a knave always lies. For instance, the model mistakenly assumes that knights lie, while knaves tell the truth. Alternatively, the model assumes that knights can only talk about other knights and knaves can only talk about knaves.)\\
    - Misunderstanding logical operators (Logical operators and their implications are not properly understood. The model fails at inferring the logical implications of a statement. For instance, if character A's statement is a logical conjunction (e.g. B and C) and A is a knave, the model does not consider all possible implications of that lie (not B and C, B and not C, not B and not C).)\\
    - Unjustified conclusion (A conclusion about the identity of a characters such as “X is a knights/knave” is presented without ANY justification. Note that this error does not address the logical validity of a justification, but checks whether a justification is present or not. As an example of this error, the model might simply state: "Since A is a knights/knave..." without prior justification.)\\
    - Unfaithfulness (A conclusion is drawn that explicitly contradicts a conclusion previously inferred by the model, without the model reflecting on this contradiction.)\\
    \vspace{2mm}
    For each error type, you must classify whether the error is present in the model's response (yes) or not (no). Start with a short justification for your decision (max. 1-2 sentences) and then clearly state whether the error type is present (yes) or not (no).\\
    Strictly adhere to the following output format:\\
    \vspace{2mm}
    \#\#\# False statement reproduction\\
    - Justification: ...\\
    - Label: yes/no\\
    \vspace{2mm}
    \#\#\# Assuming statements to be true\\
    - Justification: ...\\
    - Label: yes/no\\
    \vspace{2mm}
    \#\#\# Misunderstanding the concept of truth and lies\\
    - Justification: ...\\
    - Label: yes/no\\
    \vspace{2mm}
    ...\\
    \vspace{2mm}
    You must strictly follow this format. Do not generate any additional text.\\</|system|>.
    };
\end{tikzpicture}

%% file: tikz/appendix/example_1_error_evaluator.tex
\begin{tikzpicture}[
    font=\fontsize{6}{7.6}\selectfont,
    ]
    
    \node[rectangle, rounded corners, minimum width=0.9\textwidth, minimum height=1.5cm, fill=cyan!20, text width=0.85\textwidth, align=left] (example4) {
    <|user|>\\
    \#\#\# Instruction \#\#\#\\
    You are given a PROBLEM STATEMENT and the RESPONSE of a large language model, respectively. Your task is to classify whether a set of errors are present in the model's response.
    For each of the following error types, classify whether the error is present in the model's response (yes) or not (no).\\
    \vspace{2mm}
    - False statement reproduction (A statement of the problem description is repeated incorrectly.)\\
    - Assuming statements to be true (The possibility that a character's statement is a lie is not considered. The model does not question the premises' truth but assumes it as given.)\\
    - Misunderstanding the concept of truth and lies (The model does not understand that a truth-teller always tells the truth and a liar always lies. For instance, the model mistakenly assumes that truth-tellers lie, while liars tell the truth. Alternatively, the model assumes that truth-tellers can only talk about other truth-tellers and liars can only talk about liars.)\\
    - Misunderstanding logical operators (Logical operators and their implications are not properly understood. The model fails at inferring the logical implications of a statement. For instance, if character A's statement is a logical conjunction (e.g. B and C) and A is a liar, the model does not consider all possible implications of that lie (not B and C, B and not C, not B and not C).)\\
    - Unjustified conclusion (A conclusion about the identity of a characters such as “X is a truth-teller/liar” is presented without ANY justification. Note that this error does not address the logical validity of a justification, but checks whether a justification is present or not. As an example of this error, the model might simply state: "Since A is a truth-teller/liar..." without prior justification.)\\
    - Unfaithfulness (A conclusion is drawn that explicitly contradicts a conclusion previously inferred by the model, without the model reflecting on this contradiction.)\\
    \vspace{2mm}
    Start with a short justification for your decision (max. 1-2 sentences). Then, clearly state whether the error type is present (yes) or not (no).\\
    \vspace{2mm}
    \#\#\# PROBLEM STATEMENT\\
    A: If C is a truth-teller, then D is a truth-teller.\\
    B: C is a liar and A is a liar.\\
    C: D is a liar and B is a liar.\\
    D: B is a truth-teller and C is a liar.\\
    \vspace{2mm}
    \#\#\# RESPONSE\\
    1. A: If C is a truth-teller, then D is a truth-teller.\\
    This statement is made by A, who is either a truth-teller or a liar. If A is a truth-teller, then the statement is true, and D is indeed a truth-teller. But if A is a liar, then the statement is false, and D is not a truth-teller. So, A's statement is true only if A is a truth-teller.\\
    \vspace{2mm}
    2. B: C is a liar and A is a liar.\\
    This statement is made by B, who is either a truth-teller or a liar. If B is a truth-teller, then the statement is false, as C is not a liar. But if B is a liar, then the statement is true, as C is indeed a liar. So, B's statement is true only if B is a liar.\\
    \vspace{2mm}
    3. C: D is a liar and B is a liar.\\
    This statement is made by C, who is either a truth-teller or a liar. If C is a truth-teller, then the statement is false, as B is not a liar. But if C is a liar, then the statement is true, as D is indeed a liar. So, C's statement is true only if C is a liar.
    \vspace{2mm}
    4. D: B is a truth-teller and C is a liar.\\
    This statement is made by D, who is either a truth-teller or a liar. If D is a truth-teller, then the statement is true, as B is indeed a truth-teller. But if D is a liar, then the statement is false, as C is not a liar. So, D's statement is true only if D is a truth-teller.\\
    \vspace{2mm}
    Now, let's combine the information:\\
    * A's statement is true only if A is a truth-teller.\\
    * B's statement is true only if B is a liar.\\
    * C's statement is true only if C is a liar.\\
    * D's statement is true only if D is a truth-teller.\\
    \vspace{2mm}
    From statement 2, we know B is a liar. From statement 4, we know D is a truth-teller.\\
    From statement 1, we know that if A is a truth-teller, then D is a truth-teller. Since we already know D is a truth-teller, A must be a truth-teller.\\
    From statement 3, we know C is a liar.\\
    \vspace{2mm}
    CONCLUSION:\\
    A: truth-teller\\
    B: liar\\
    C: liar\\
    D: truth-teller\\
    </|user|>
    };
    \node[rectangle, rounded corners, minimum width=0.9\textwidth, minimum height=1.0cm, fill=orange!20, text width=0.85\textwidth, align=left, below=0.0\textwidth of example4.south, anchor=north] (answer1) {
    <|assistant|>\\
    \#\#\# False statement reproduction\\
    - Justification: The statements are properly reproduced by the model.\\
    - Label: no\\
    \vspace{2mm}
    \#\#\# Assuming statements to be true\\
    - Justification: The possibility that the characters' statements are lies is considered by the model.\\
    - Label: no\\
    \vspace{2mm}
    \#\#\# Misunderstanding the concept of truth and lies\\
    - Justification: When discussing B's statement, the model says: "So, B's statement is true only if B is a liar". However, if B were a liar, its statement would not be true. This shows a model failure to understand the concept of a liar.\\
    - Label: yes\\
    \vspace{2mm}
    \#\#\# Misunderstanding logical operators\\
    - Justification: The model fails to understand the logical implications of A's statement: "If C is a truth-teller, then D is a truth-teller." It draws conclusions about D based on the identity of A, ignoring character C, who is the antecedent of the logical implication.\\
    - Label: yes\\
    \vspace{2mm}
    \#\#\# Unjustified conclusion\\
    - Justification: When discussing B's statement, the model defines the identity of character C: "as C is indeed a liar", without proper justification.\\
    - Label: yes\\
    \vspace{2mm}
    \#\#\# Unfaithfulness\\
    - Justification: The model does not draw any conclusions that contradict previous conclusions.\\
    - Label: no\\
    </|assistant|>
    };
\end{tikzpicture}

%% file: tikz/appendix/example_2_error_evaluator.tex
\begin{tikzpicture}[
    font=\fontsize{6}{7.6}\selectfont,
    ]
    
    \node[rectangle, rounded corners, minimum width=0.9\textwidth, minimum height=1.5cm, fill=cyan!20, text width=0.85\textwidth, align=left] (example4) {
    <|user|>\\
    \#\#\# Instruction \#\#\#\\
    You are given a PROBLEM STATEMENT and the RESPONSE of a large language model, respectively. Your task is to classify whether a set of errors are present in the model's response.
    For each of the following error types, classify whether the error is present in the model's response (yes) or not (no).\\
    \vspace{2mm}
    - False statement reproduction (A statement of the problem description is repeated incorrectly.)\\
    - Assuming statements to be true (The possibility that a character's statement is a lie is not considered. The model does not question the premises' truth but assumes it as given.)\\
    - Misunderstanding the concept of truth and lies (The model does not understand that a truth-teller always tells the truth and a liar always lies. For instance, the model mistakenly assumes that truth-tellers lie, while liars tell the truth. Alternatively, the model assumes that truth-tellers can only talk about other truth-tellers and liars can only talk about liars.)\\
    - Misunderstanding logical operators (Logical operators and their implications are not properly understood. The model fails at inferring the logical implications of a statement. For instance, if character A's statement is a logical conjunction (e.g. B and C) and A is a liar, the model does not consider all possible implications of that lie (not B and C, B and not C, not B and not C).)\\
    - Unjustified conclusion (A conclusion about the identity of a characters such as “X is a truth-teller/liar” is presented without ANY justification. Note that this error does not address the logical validity of a justification, but checks whether a justification is present or not. As an example of this error, the model might simply state: "Since A is a truth-teller/liar..." without prior justification.)\\
    - Unfaithfulness (A conclusion is drawn that explicitly contradicts a conclusion previously inferred by the model, without the model reflecting on this contradiction.)\\
    \vspace{2mm}
    Start with a short justification for your decision (max. 1-2 sentences). Then, clearly state whether the error type is present (yes) or not (no).\\
    \vspace{2mm}
    \#\#\# PROBLEM STATEMENT\\
    A: C is a liar and B is a truth-teller.\\
    B: I am a truth-teller.\\
    C: A is a truth-teller.\\
    \vspace{2mm}
    \#\#\# RESPONSE\\
    Great, let's dive into the problem!\\
    To solve this problem, we need to use logical reasoning and deduction. We are given three statements from three different characters, and our goal is to determine who is a truth-teller and who is a liar based on their statements.\\
    Statement A tells us that C is a liar and B is a truth-teller. This means that B is telling the truth, since he is a truth-teller.\\
    Statement B tells us that I (B) am a truth-teller. This means that I am telling the truth, since I am a truth-teller.\\
    Statement C tells us that A is a truth-teller. This means that A is telling the truth, since A is a truth-teller.\\
    Now, let's use these statements to deduce the identity of each character:\\
    \vspace{2mm}
    A: truth-teller\\
    B: truth-teller\\
    C: truth-teller\\
    </|user|>
    };
    \node[rectangle, rounded corners, minimum width=0.9\textwidth, minimum height=1.0cm, fill=orange!20, text width=0.85\textwidth, align=left, below=0.0\textwidth of example4.south, anchor=north] (answer1) {
    <|assistant|>\\
    \#\#\# False statement reproduction\\
    - Justification: No occurrence of this error in the model's response.\\
    - Label: no\\
    \vspace{2mm}
    \#\#\# Assuming statements to be true\\
    - Justification: The model treats each character's statement as a true premise and disregards the possibility that statements might be lies. See for instance: "A tells us that C is a liar and B is a truth-teller. This means that B is telling the truth, since he is a truth-teller."\\
    - Label: yes\\
    \vspace{2mm}
    \#\#\# Misunderstanding the concept of truth and lies\\
    - Justification: No occurrence of this error in the model's response.\\
    - Label: no\\
    \vspace{2mm}
    \#\#\# Misunderstanding logical operators\\
    - Justification: No occurrence of this error in the model's response.\\
    - Label: no\\
    \vspace{2mm}
    \#\#\# Unjustified conclusion\\
    - Justification: No occurrence of this error in the model's response.\\
    - Label: no\\
    \vspace{2mm}
    \#\#\# Unfaithfulness\\
    - Justification: No occurrence of this error in the model's response.\\
    - Label: no\\
    </|assistant|>
    };
\end{tikzpicture}

%% file: tikz/appendix/example_3_error_evaluator.tex
\begin{tikzpicture}[
    font=\fontsize{6}{7.6}\selectfont,
    ]
    
    \node[rectangle, rounded corners, minimum width=0.9\textwidth, minimum height=1.5cm, fill=cyan!20, text width=0.85\textwidth, align=left] (example4) {
    <|user|>\\
    \#\#\# Instruction \#\#\#\\
    You are given a PROBLEM STATEMENT and the RESPONSE of a large language model, respectively. Your task is to classify whether a set of errors are present in the model's response.
    For each of the following error types, classify whether the error is present in the model's response (yes) or not (no).\\
    \vspace{2mm}
    - False statement reproduction (A statement of the problem description is repeated incorrectly.)\\
    - Assuming statements to be true (The possibility that a character's statement is a lie is not considered. The model does not question the premises' truth but assumes it as given.)\\
    - Misunderstanding the concept of truth and lies (The model does not understand that a truth-teller always tells the truth and a liar always lies. For instance, the model mistakenly assumes that truth-tellers lie, while liars tell the truth. Alternatively, the model assumes that truth-tellers can only talk about other truth-tellers and liars can only talk about liars.)\\
    - Misunderstanding logical operators (Logical operators and their implications are not properly understood. The model fails at inferring the logical implications of a statement. For instance, if character A's statement is a logical conjunction (e.g. B and C) and A is a liar, the model does not consider all possible implications of that lie (not B and C, B and not C, not B and not C).)\\
    - Unjustified conclusion (A conclusion about the identity of a characters such as “X is a truth-teller/liar” is presented without ANY justification. Note that this error does not address the logical validity of a justification, but checks whether a justification is present or not. As an example of this error, the model might simply state: "Since A is a truth-teller/liar..." without prior justification.)\\
    - Unfaithfulness (A conclusion is drawn that explicitly contradicts a conclusion previously inferred by the model, without the model reflecting on this contradiction.)\\
    \vspace{2mm}
    Start with a short justification for your decision (max. 1-2 sentences). Then, clearly state whether the error type is present (yes) or not (no).\\
    \vspace{2mm}
    \#\#\# PROBLEM STATEMENT\\
    A: C is a truth-teller and B is a liar.\\
    B: If A is a truth-teller, then C is a truth-teller.\\
    C: B is a liar and A is a liar.\\
    \vspace{2mm}
    \#\#\# RESPONSE\\
    REASONING:\\
    Let's analyze the statements:\\
    \vspace{2mm}
    A: C is a truth-teller and B is a liar.\\
    B: If A is a truth-teller, then C is a truth-teller.\\
    C: B is a liar and A is a liar.\\
    \vspace{2mm}
    If A is a truth-teller, then A's statement must be true, so C is a truth-teller and B is a liar.\\
    If C is a truth-teller, then C's statement must be false, so either B is not a liar or A is not a liar.\\
    This contradicts our assumption that A is a truth-teller, so A must be a liar.\\
    \vspace{2mm}
    If A is a liar, then A's statement must be false, so either C is not a truth-teller or B is not a liar.\\
    Since A's statement is false, we can't conclude anything about C, but we know B is not a liar.\\
    If B is a truth-teller, then B's statement must be true, so A is a liar and C is a truth-teller.\\
    This is consistent with our previous conclusion that A is a liar.\\
    If C is a truth-teller, then C's statement must be false, so either B is not a liar or A is not a liar.\\
    We know A is a liar, so B must be a truth-teller.\\
    \vspace{2mm}
    CONCLUSION:\\
    A: liar\\
    B: truth-teller\\
    C: truth-teller\\
    </|user|>
    };
    \node[rectangle, rounded corners, minimum width=0.9\textwidth, minimum height=1.0cm, fill=orange!20, text width=0.85\textwidth, align=left, below=0.0\textwidth of example4.south, anchor=north] (answer1) {
    <|assistant|>\\
    \#\#\# False statement reproduction\\
    - Justification: The model correctly repeats the statements in the problem description.\\
    - Label: no\\
    \vspace{2mm}
    \#\#\# Assuming statements to be true\\
    - Justification: The model considers the possibility that statements might be false.\\
    - Label: no\\
    \vspace{2mm}
    \#\#\# Misunderstanding the concept of truth and lies\\
    - Justification: When discussing C's statement, the model says: "If C is a truth-teller, then C's statement must be false". However, if C were a truth-teller, its statement could not be false. This shows a failure of the model to understand the concept of truth.\\
    - Label: yes\\
    \vspace{2mm}
    \#\#\# Misunderstanding logical operators\\
    - Justification: The model exhibits several problems to draw the logical implications of statements. For example, it says: "Since A's statement is false, we can't conclude anything about C". However, the logical conjunction of A's statement allows us to draw conclusions about C.\\
    - Label: yes\\
    \vspace{2mm}
    \#\#\# Unjustified conclusion\\
    - Justification: No occurrence of this error in the model's response.\\
    - Label: no\\
    \vspace{2mm}
    \#\#\# Unfaithfulness\\
    - Justification: No occurrence of this error in the model's response.\\
    - Label: no\\
    </|assistant|>
    };
\end{tikzpicture}

%% file: tikz/appendix/example_4_error_evaluator.tex
\begin{tikzpicture}[
    font=\fontsize{6}{7.6}\selectfont,
    ]
    
    \node[rectangle, rounded corners, minimum width=0.9\textwidth, minimum height=1.5cm, fill=cyan!20, text width=0.85\textwidth, align=left] (example4) {
    <|user|>\\
    \#\#\# Instruction \#\#\#\\
    You are given a PROBLEM STATEMENT and the RESPONSE of a large language model, respectively. Your task is to classify whether a set of errors are present in the model's response.
    For each of the following error types, classify whether the error is present in the model's response (yes) or not (no).\\
    \vspace{2mm}
    - False statement reproduction (A statement of the problem description is repeated incorrectly.)\\
    - Assuming statements to be true (The possibility that a character's statement is a lie is not considered. The model does not question the premises' truth but assumes it as given.)\\
    - Misunderstanding the concept of truth and lies (The model does not understand that a truth-teller always tells the truth and a liar always lies. For instance, the model mistakenly assumes that truth-tellers lie, while liars tell the truth. Alternatively, the model assumes that truth-tellers can only talk about other truth-tellers and liars can only talk about liars.)\\
    - Misunderstanding logical operators (Logical operators and their implications are not properly understood. The model fails at inferring the logical implications of a statement. For instance, if character A's statement is a logical conjunction (e.g. B and C) and A is a liar, the model does not consider all possible implications of that lie (not B and C, B and not C, not B and not C).)\\
    - Unjustified conclusion (A conclusion about the identity of a characters such as “X is a truth-teller/liar” is presented without ANY justification. Note that this error does not address the logical validity of a justification, but checks whether a justification is present or not. As an example of this error, the model might simply state: "Since A is a truth-teller/liar..." without prior justification.)\\
    - Unfaithfulness (A conclusion is drawn that explicitly contradicts a conclusion previously inferred by the model, without the model reflecting on this contradiction.)\\
    \vspace{2mm}
    Start with a short justification for your decision (max. 1-2 sentences). Then, clearly state whether the error type is present (yes) or not (no).\\
    \vspace{2mm}
    \#\#\# PROBLEM STATEMENT\\
    Assume that there exist only two types of people: truth-tellers and liars. truth-tellers always tell the truth, while liars always lie.\\
    You are given the statements from a set of characters. Based on their statements, infer who is a truth-teller and who is a liar:\\
    A: D is a truth-teller.\\
    B: C is a liar and A is a liar.\\
    C: A is a liar.\\
    D: A is a liar and B is a truth-teller.\\
    \vspace{2mm}
    \#\#\# RESPONSE\\
    Let's analyze the statements given:\\
    1. A: D is a truth-teller.\\
    This statement is made by A, who according to statement D, is a liar. Therefore, this statement is a lie, and D is not a truth-teller.\\
    2. B: C is a liar and A is a liar.\\
    This statement is made by B, who according to statement D, is a truth-teller. Therefore, this statement is true, and C and A are both liars.\\
    3. C: A is a liar.\\
    This statement is made by C, who according to statement B, is a liar. Therefore, this statement is true, and A is a liar.\\
    4. D: A is a liar and B is a truth-teller.\\
    This statement is made by D, who according to statement A, is a truth-teller. Therefore, this statement is true, and A is a liar and B is a truth-teller.\\
    \vspace{2mm}
    Now, let's summarize our findings:\\
    * A is a liar (according to statements B and C).\\
    * B is a truth-teller (according to statement D).\\
    * C is a liar (according to statement B).\\
    * D is a truth-teller (according to statement A).\\
    \vspace{2mm}
    CONCLUSION:\\
    A: truth-teller\\
    B: liar\\
    C: liar\\
    D: truth-teller\\
    </|user|>
    };
    \node[rectangle, rounded corners, minimum width=0.9\textwidth, minimum height=1.0cm, fill=orange!20, text width=0.85\textwidth, align=left, below=0.0\textwidth of example4.south, anchor=north] (answer1) {
    <|assistant|>\\
    \#\#\# False statement reproduction\\
    - Justification: The model correctly repeats the statements in the problem description.\\
    - Label: no\\
    \vspace{2mm}
    \#\#\# Assuming statements to be true\\
    - Justification: The model considers the possibility that statements might be false.\\
    - Label: no\\
    \vspace{2mm}
    \#\#\# Misunderstanding the concept of truth and lies\\
    - Justification: When discussing C's statement, the model says: "This statement is made by C, who according to statement B, is a liar. Therefore, this statement is true...". However, if C were a liar, its statement could not be true. This shows a failure of the model to understand the concept of truth.\\
    - Label: yes\\
    \vspace{2mm}
    \#\#\# Misunderstanding logical operators\\
    - Justification: The model does not show any failures at understanding logical operators.\\
    - Label: no\\
    \vspace{2mm}
    \#\#\# Unjustified conclusion\\
    - Justification: When discussion the implications of A's statement, the model defines D's statement as truth without properly justifying this assumption: "This statement is made by A, who according to statement D, is a liar. Therefore, this statement is a lie, and D is not a truth-teller."\\
    - Label: yes\\
    \vspace{2mm}
    \#\#\# Unfaithfulness\\
    - Justification: The model first concludes that "D is not a truth-teller", but then follows up "This statement is made by D, who according to statement A, is a truth-teller. Therefore, this statement is true." This contradicts the initial conclusion.\\
    - Label: yes\\
    </|assistant|>
    };
\end{tikzpicture}

%% file: tikz/appendix/example_5_error_evaluator.tex
\begin{tikzpicture}[
    font=\fontsize{6}{7.6}\selectfont,
    ]
    
    \node[rectangle, rounded corners, minimum width=0.9\textwidth, minimum height=1.5cm, fill=cyan!20, text width=0.85\textwidth, align=left] (example4) {
    <|user|>\\
    \#\#\# Instruction \#\#\#\\
    You are given a PROBLEM STATEMENT and the RESPONSE of a large language model, respectively. Your task is to classify whether a set of errors are present in the model's response.
    For each of the following error types, classify whether the error is present in the model's response (yes) or not (no).\\
    \vspace{2mm}
    - False statement reproduction (A statement of the problem description is repeated incorrectly.)\\
    - Assuming statements to be true (The possibility that a character's statement is a lie is not considered. The model does not question the premises' truth but assumes it as given.)\\
    - Misunderstanding the concept of truth and lies (The model does not understand that a truth-teller always tells the truth and a liar always lies. For instance, the model mistakenly assumes that truth-tellers lie, while liars tell the truth. Alternatively, the model assumes that truth-tellers can only talk about other truth-tellers and liars can only talk about liars.)\\
    - Misunderstanding logical operators (Logical operators and their implications are not properly understood. The model fails at inferring the logical implications of a statement. For instance, if character A's statement is a logical conjunction (e.g. B and C) and A is a liar, the model does not consider all possible implications of that lie (not B and C, B and not C, not B and not C).)\\
    - Unjustified conclusion (A conclusion about the identity of a characters such as “X is a truth-teller/liar” is presented without ANY justification. Note that this error does not address the logical validity of a justification, but checks whether a justification is present or not. As an example of this error, the model might simply state: "Since A is a truth-teller/liar..." without prior justification.)\\
    - Unfaithfulness (A conclusion is drawn that explicitly contradicts a conclusion previously inferred by the model, without the model reflecting on this contradiction.)\\
    \vspace{2mm}
    Start with a short justification for your decision (max. 1-2 sentences). Then, clearly state whether the error type is present (yes) or not (no).\\
    \vspace{2mm}
    \#\#\# PROBLEM STATEMENT\\
    A: F is a liar.\\
    B: A is a truth-teller.\\
    C: D is a truth-teller and B is a truth-teller.\\
    D: E is a truth-teller.\\
    E: D is a liar and B is a truth-teller.\\
    F: I am a truth-teller.\\
    \vspace{2mm}
    \#\#\# RESPONSE\\
    Let's start by analyzing the statements:\\
    \vspace{2mm}
    1. A: F is a liar.\\
    If A is a truth-teller, then F is indeed a liar, which means A is a truth-teller. But if A is a liar, then F is not a liar, which is a contradiction. Therefore, A must be a truth-teller.\\
    2. B: A is a truth-teller.\\
    Since A is a truth-teller, this statement is true. Therefore, B is a truth-teller.\\
    3. C: D is a truth-teller and B is a truth-teller.\\
    Since C is a truth-teller, this statement is true. Therefore, D is a truth-teller, and B is indeed a truth-teller.\\
    4. D: E is a truth-teller.\\
    Since D is a truth-teller, this statement is true. Therefore, E is a truth-teller.\\
    5. E: D is a liar and B is a truth-teller.\\
    Since E is a liar, this statement is false. However, we know that D is a truth-teller from statement 3, and B is a truth-teller from statements 2 and 3. Therefore, E must be a truth-teller.\\
    6. F: I am a truth-teller.\\
    Since F is a liar, this statement is false. Therefore, F is a liar.\\
    \vspace{2mm}
    Now, let's summarize the identities:\\
    \vspace{2mm}
    A: truth-teller\\
    B: truth-teller\\
    C: truth-teller\\
    D: truth-teller\\
    E: truth-teller\\
    F: liar\\
    </|user|>
    };
    \node[rectangle, rounded corners, minimum width=0.9\textwidth, minimum height=1.0cm, fill=orange!20, text width=0.85\textwidth, align=left, below=0.0\textwidth of example4.south, anchor=north] (answer1) {
    <|assistant|>\\
    \#\#\# False statement reproduction\\
    - Justification: The model correctly repeats the statements in the problem description.\\
    - Label: no\\
    \vspace{2mm}
    \#\#\# Assuming statements to be true\\
    - Justification: The model considers the possibility that statements might be false.\\
    - Label: no\\
    \vspace{2mm}
    \#\#\# Misunderstanding the concept of truth and lies\\
    - Justification: The model's response does not exhibit this type of error.\\
    - Label: no\\
    \vspace{2mm}
    \#\#\# Misunderstanding logical operators\\
    - Justification: The model does not show any failures at understanding logical operators.\\
    - Label: no\\
    \vspace{2mm}
    \#\#\# Unjustified conclusion\\
    - Justification: When discussion C's statement, the model defines C as a truth-teller without proper prior justification.\\
    - Label: yes\\
    \vspace{2mm}
    \#\#\# Unfaithfulness\\
    - Justification: The model first concludes that "E is a truth-teller". However, it then continues with: "Since E is a liar, this statement is false", which contradicts the previous conclusion.\\
    - Label: yes\\
    </|assistant|>
    };
\end{tikzpicture}

%% file: tikz/appendix/example_6_error_evaluator.tex
\begin{tikzpicture}[
    font=\fontsize{6}{7.6}\selectfont,
    ]
    
    \node[rectangle, rounded corners, minimum width=0.9\textwidth, minimum height=1.5cm, fill=cyan!20, text width=0.85\textwidth, align=left] (example4) {
    <|user|>\\
    \#\#\# Instruction \#\#\#\\
    You are given a PROBLEM STATEMENT and the RESPONSE of a large language model, respectively. Your task is to classify whether a set of errors are present in the model's response.
    For each of the following error types, classify whether the error is present in the model's response (yes) or not (no).\\
    \vspace{2mm}
    - False statement reproduction (A statement of the problem description is repeated incorrectly.)\\
    - Assuming statements to be true (The possibility that a character's statement is a lie is not considered. The model does not question the premises' truth but assumes it as given.)\\
    - Misunderstanding the concept of truth and lies (The model does not understand that a truth-teller always tells the truth and a liar always lies. For instance, the model mistakenly assumes that truth-tellers lie, while liars tell the truth. Alternatively, the model assumes that truth-tellers can only talk about other truth-tellers and liars can only talk about liars.)\\
    - Misunderstanding logical operators (Logical operators and their implications are not properly understood. The model fails at inferring the logical implications of a statement. For instance, if character A's statement is a logical conjunction (e.g. B and C) and A is a liar, the model does not consider all possible implications of that lie (not B and C, B and not C, not B and not C).)\\
    - Unjustified conclusion (A conclusion about the identity of a characters such as “X is a truth-teller/liar” is presented without ANY justification. Note that this error does not address the logical validity of a justification, but checks whether a justification is present or not. As an example of this error, the model might simply state: "Since A is a truth-teller/liar..." without prior justification.)\\
    - Unfaithfulness (A conclusion is drawn that explicitly contradicts a conclusion previously inferred by the model, without the model reflecting on this contradiction.)\\
    \vspace{2mm}
    Start with a short justification for your decision (max. 1-2 sentences). Then, clearly state whether the error type is present (yes) or not (no).\\
    \vspace{2mm}
    \#\#\# PROBLEM STATEMENT\\
    A: B is a liar if and only if C is a liar.\\
    B: C is a liar.\\
    C: A is a liar.\\
    \vspace{2mm}
    \#\#\# RESPONSE\\
    REASONING:\\
    Let's assume B is a truth-teller.\\
    If B is a truth-teller, B's statement is true and C is a liar.\\
    If C is a liar, C's statement is false and A is a truth-teller.\\
    If A is a truth-teller, A's statement is true, so either B is a liar and C is a liar or B is a truth-teller and C is a truth-teller.\\
    Since we know C is a liar, B must be a liar.\\
    This leads to a contradiction, so B must be a liar.\\
    If B is a liar, B's statement is false and C is a truth-teller.\\
    If C is a truth-teller, C's statement is true and A is a liar.\\
    If A is a liar, A's statement is false, so either B is a truth-teller and C is a truth-teller or B is a liar and C is a liar.\\
    Since we know B is a liar, C must be a truth-teller.\\
    Therefore, we can conclude:\\
    \vspace{2mm}
    CONCLUSION:\\
    A: liar\\
    B: liar\\
    C: truth-teller\\
    </|user|>
    };
    \node[rectangle, rounded corners, minimum width=0.9\textwidth, minimum height=1.0cm, fill=orange!20, text width=0.85\textwidth, align=left, below=0.0\textwidth of example4.south, anchor=north] (answer1) {
    <|assistant|>\\
    \#\#\# False statement reproduction\\
    - Justification: The model correctly repeats the statements in the problem description.\\
    - Label: no\\
    \vspace{2mm}
    \#\#\# Assuming statements to be true\\
    - Justification: The model considers the possibility that statements might be false.\\
    - Label: no\\
    \vspace{2mm}
    \#\#\# Misunderstanding the concept of truth and lies\\
    - Justification: The model seem to capture the concept of truth and lies.\\
    - Label: no\\
    \vspace{2mm}
    \#\#\# Misunderstanding logical operators\\
    - Justification: The model fails to infer the logical implications of A's false statement. Although the model identifies A's statement as a lie, it does not successfully build the negation of the logical equivalence: "If A is a liar, A's statement is false, so either B is a truth-teller and C is a truth-teller or B is a liar and C is a liar."\\
    - Label: yes\\
    \vspace{2mm}
    \#\#\# Unjustified conclusion\\
    - Justification: All conclusions are justified.\\
    - Label: no\\
    \vspace{2mm}
    \#\#\# Unfaithfulness\\
    - Justification: The model does not infer conclusions that contradict conclusions previously drawn.\\
    - Label: no\\
    </|assistant|>
    };
\end{tikzpicture}